\documentclass[11pt]{article}

\usepackage{tikz}
\usetikzlibrary{shapes.geometric, arrows}
\tikzstyle{arrow} = [thick,->,>=stealth]

\usepackage{graphicx}
\usepackage{enumitem}

\usepackage{amsmath}
\usepackage[showonlyrefs]{mathtools}
\usepackage{mathstyle}
\usepackage{amstext,amssymb,amsfonts}
\usepackage{fullpage}
\usepackage{nicefrac}
\usepackage{bm}
\usepackage{textcomp}
\usepackage{mdwlist}
\usepackage[colorinlistoftodos]{todonotes}
\usepackage{subcaption}

\usepackage[linesnumbered,ruled]{algorithm2e}

\usepackage[pagebackref,colorlinks,linkcolor=blue,filecolor = blue,citecolor = blue, urlcolor = blue, hyperfootnotes=false]{hyperref}
\usepackage[capitalize, noabbrev, nameinlink]{cleveref}

\usepackage{amsthm}
\usepackage{thmtools}

\declaretheorem[within=section]{theorem}
\declaretheorem[sibling=theorem]{corollary}

\declaretheorem[sibling=theorem]{lemma}
\declaretheorem[sibling=theorem]{claim}

\declaretheorem[sibling=theorem]{definition}

\declaretheorem[sibling=theorem]{assumption}

\crefname{assumption}{Assumption}{Assumptions}

\newcounter{termcounter}
\renewcommand{\thetermcounter}{\Alph{termcounter}}
\crefname{term}{term}{terms}
\creflabelformat{term}{\textup{(#2#1#3)}}

\makeatletter
\def\term{\@ifnextchar[\term@optarg\term@noarg}
\def\term@optarg[#1]#2{%
  \textup{(#1)}%
  \def\@currentlabel{#1}%
  \def\cref@currentlabel{[][2147483647][]#1}%
  \cref@label[term]{#2}}
\def\term@noarg#1{%
  \refstepcounter{termcounter}%
  \textup{(\thetermcounter)}%
  \cref@label[term]{#1}}
\makeatother

\newcommand{\mv}[1]{\mathbf {#1}}

\newcommand{\ignore}[1]{}

\newcommand{\defeq}{\stackrel{\mathrm{def}}=}

\newcommand{\brac}[1]{[#1 ]}

\renewcommand{\vec}[1]{{\bf{#1}}}

\renewcommand{\leq}{\leqslant}
\renewcommand{\geq}{\geqslant}
\renewcommand{\ge}{\geqslant}
\renewcommand{\le}{\leqslant}
\renewcommand{\epsilon}{\varepsilon}
\newcommand{\eps}{\epsilon}

\newcommand{\cP}{\mathcal P}
\newcommand{\cQ}{\mathcal Q}

\newcommand{\Esymb}{{\bf E}}

\newcommand{\Psymb}{{\bf Pr}}

\DeclareMathOperator*{\E}{\Esymb}

\DeclareMathOperator*{\ProbOp}{\Psymb}

\renewcommand{\Pr}{\ProbOp}

\newcommand{\indep}{\bot}

\newcommand{\ex}[1]{\E\brac{#1}}

\newcommand{\dtv}{\mbox{${d}_{\mathrm{TV}}$}}

\newcommand{\dkl}{\mathrm{KL}}

\DeclareMathOperator*{\Pa}{\bm{\mathsf{Pa}}}
\DeclareMathOperator*{\pa}{\bm{\mathsf{pa}}}
\DeclareMathOperator*{\An}{\bm{\mathsf{An}}}
\DeclareMathOperator*{\an}{\bm{\mathsf{an}}}
\DeclareMathOperator*{\De}{\bm{\mathsf{De}}}
\DeclareMathOperator*{\de}{\bm{\mathsf{de}}}

\newcommand{\cdo}[0]{\mathsf{do}}
\newcommand{\bern}[0]{\mathrm{Bern}}

\def\notes{1}
 \newcommand{\anote}[1]{\ifnum\notes=1{\mnote{Arnab: #1}}\fi}
 \newcommand{\snote}[1]{\ifnum\notes=1{\mnote{Sutanu: #1}}\fi}
 \newcommand{\vnote}[1]{\ifnum\notes=1{\mnote{Vinod: #1}}\fi}
\newcommand{\knote}[1]{\ifnum\notes=1{\mnote{Saravanan: #1}}\fi}

\begin{document}

\title{Learning and Sampling of Atomic Interventions from Observations\footnote{Author names are in the alphabetical order.}}

\author{
Arnab Bhattacharyya\thanks{National University of Singapore. Supported in part by Start-up Grant WBS R252000A33133.}\\
arnabb@nus.edu.sg
 \and 
Sutanu Gayen\thanks{National University of Singapore. Supported in part by AB's Start-up Grant WBS R252000A33133.}\\
sutanugayen@gmail.com
\and 
Saravanan Kandasamy\thanks{Cornell University.  Supported by Cornell University Grant.}\\
sk3277@cornell.edu
\and 
Ashwin Maran\thanks{University of Wisconsin-Madison.}\\
amaran@wisc.edu
\and 
N.~V.~Vinodchandran\thanks{University of Nebraska-Lincoln. Research mostly conducted while visiting National University of Singapore.}\\
vinod@cse.unl.edu
}





\maketitle
\begin{abstract}
We study the problem of efficiently estimating the effect of an intervention on a single variable (atomic interventions) using observational samples in a causal Bayesian network. Our goal is to give algorithms that are efficient in both  time and sample complexity in a non-parametric setting.

Tian and Pearl (AAAI `02) have exactly characterized the class of causal graphs for which causal effects of atomic interventions can be identified from observational data. We make their result quantitative.  Suppose $\mathcal{P}$ is a causal model on a set $\vec{V}$ of $n$ observable variables with respect to a given causal graph $G$ with observable distribution $P$.  Let $P_x$ denote the interventional distribution over the observables with respect to an intervention of a designated variable $X$ with $x$.\footnote{$P(\vec{V} \mid \cdo(x))$ is another notation for $P_x$ that is widely used in the literature, with $\cdo(x)$ denoting an intervention on a variable $X$ with value $x$.}  Assuming that $G$ has bounded in-degree, bounded c-components ($k$), and that the observational distribution is identifiable and satisfies certain strong positivity condition, we give an algorithm that takes $m=\tilde{O}(n\eps^{-2})$ samples from $P$ and $O(mn)$ time, and outputs with high probability a description of a distribution $\hat{P}$ such that $\dtv(P_x, \hat{P}) \leq \eps$, and:
\begin{enumerate}
\item[(i)] [Evaluation]~the description can return in $O(n)$ time the probability $\hat{P}(\vec{v})$ for any assignment $\vec{v}$ to $\vec{V}$. 
\item[(ii)][Generation]~the description can return an iid sample from $\hat{P}$ in  $O(n)$ time. 
\end{enumerate}
We extend our techniques to estimate marginals $P_x|_{\vec{Y}}$ over a given subset $\vec{Y} \subseteq \vec{V}$ of variables of interest. We also show lower bounds for the sample complexity showing that our sample complexity has an optimal dependence on the parameters $n$ and $\eps$, as well as if $k=1$ on the strong positivity parameter.
\end{abstract}

\newpage

\section{Introduction}

A causal model for a system of variables describes not only how the variables are associated with each other but also how they would change if they were to be acted on by an external force. For example, in order to have a proper discussion about global warming, we need more than just an associational model which would give the correlation between human CO$_2$ emissions and  Arctic temperature levels. We instead need a causal model which would predict the climatological effects of humans reducing CO$_2$ emissions by (say) 20$\%$ over the next five years. Notice how the two can give starkly different pictures: if global warming is being propelled by natural weather cycles, then changing human emissions won't make any difference to temperature levels, even though human emissions and temperature may be correlated in our dataset (just because both are increasing over the timespan of our data).

Causality has been a topic of inquiry since ancient times, but a modern, rigorous formulation of causality came about in the twentieth century through the works of Pearl, Robins, Rubin, and others \cite{IR15,pearl00,rubin2011,hernan-book}. In particular, Pearl \cite{pearl00} recasted causality in the language of {\em causal Bayesian networks} (or {\em causal Bayes nets} for short). A causal Bayes net is a standard Bayes net that is reinterpreted causally. Specifically, it makes the assumption of {\em modularity}:  for any variable $X$, the dependence of $X$ on its parents is an autonomous mechanism that does not change even if other
parts of the network are changed. This allows assessment of external interventions, such as those encountered in policy analysis, treatment management, and planning. The idea is that by virtue of the modularity assumption, an intervention simply amounts to a modified Bayes net where some of the parent-child mechanisms are altered while the rest are kept the same.

The underlying structure of causal Bayes net $\cP$ is a directed acyclic graph $G$. The graph $G$ consists of $n+h$ nodes where $n$ nodes correspond to the {\em observable} variables $\vec{V}$ while the $h$ additional nodes correspond to a set of $h$ {\em hidden variables} $\vec{U}$. We assume that the observable variables take values over a finite alphabet $\Sigma$. 
By interpreting $\cP$ as a standard Bayes net over $\vec{V} \cup \vec{U}$ and then marginalizing to $\vec{V}$, we get the observational distribution $P$ on $\vec{V}$. The modularity assumption allows us to define the result of an {\em intervention} on $\cP$. An intervention is specified by a subset $\vec{X} \subseteq \vec{V}$ of variables and an assignment\footnote{Consistent with the convention in the causality literature, we will use a lower case letter (e.g., $\vec{x}$) to denote an assignment to the subset of variables corresponding to its upper case counterpart (e.g., $\vec{X}$).} $\vec{x} \in \Sigma^{|\vec{X}|}$. In the  interventional distribution, the variables $\vec{X}$ are fixed to $\vec{x}$, while each variable $W \in (\vec{V} \cup \vec{U})\setminus \vec{X}$ is sampled as it would have been in the original Bayes net, according to the conditional distribution $W \mid \Pa(W)$, where $\Pa(W)$ (parents of $W$) consist of either variables previously sampled in the topological order of $G$ or variables in $\vec{X}$ set by the intervention. The marginal of the resulting distribution to $\vec{V}$ is the interventional distribution denoted by $P_{\vec{x}}$.  We sometimes also use $\cdo(\vec{x})$ to denote the intervention process and $P(\vec{V}\mid \cdo(\vec{x}))$ to denote the resulting interventional distribution.

In this work, we focus our attention on the case that $X$ is a single observable variable, so that interventions on $X$ are {\em atomic}. 
We study the following estimation problems:
\begin{enumerate}
\item[1.]\textbf{(Evaluation)} Given an $x \in \Sigma$, construct an {\em evaluator} for $P_x$ which estimates the value of the probability mass function
\[P_x(\vec{v}) \defeq \Pr_{\vec{V} \sim P_x}[\vec{V} = \vec{v}]\]
for any $\vec{v} \in \Sigma^n$. The goal is to construct the evaluator using only a bounded number of samples from the observational distribution $P$, and moreover, the evaluator should run efficiently. 
\item[2.]\textbf{(Generation)}
Given an $x \in \Sigma$, construct a {\em generator} for $P_x$ which generates i.i.d.~samples from a distribution that approximates $P_x$. The goal is to construct the generator using only a bounded number of samples from the observational distribution $P$, and moreover, the generator should be able to output each sample efficiently.
\end{enumerate}
We study these problems in the non-parametric setting, where we assume that all the observable variables under consideration are over a finite alphabet $\Sigma$.  

Evaluation and generation are two very natural inference problems\footnote{Note that the distinction between the two problems is computational; one can produce a generator from an evaluator and vice versa without requiring any new samples.}. Indeed, the influential work of Kearns et al.~\cite{KMRRSS94} introduced the computational framework of {\em distribution learning} in terms of these two problems. Over the last 25 years, work on distribution learning has clarified how classical techniques in statistics can be married to new algorithmic ideas in order to yield sample- and time-efficient algorithms for learning very general classes of distributions; see \cite{D16} for a recent survey of the area. The goal of our work is to initiate a similar computational study of the fundamental problems in causal inference. 

The crucial distinction of our setting from the distribution learning setting is that the algorithm {\em does not get samples from the distribution of interest}. In our setting, the algorithm receives as input samples from $P$ while its goal is to estimate the distribution $P_x$. This is motivated by the fact that typically randomized experiments are hard (or unethical) to conduct while observational samples are easy to collect.  Even if we disregard computational considerations, it may be impossible to determine the interventional distribution $P_x$ from the observational distribution $P$ and knowledge of the causal graph $G$. The simplest example is the so-called ``bow-tie graph'' on two observable variables $X$ and $Y$ (with $X$ being a parent of $Y$) and a hidden variable $U$ that is a parent of both $X$ and $Y$. Here, it's easy to see that $P$ does not uniquely determine $P_x$.  Tian and Pearl~\cite{TP02b} studied the general question of when the interventional distribution $P_x$ is identifiable from the observational distribution $P$. They characterized the class $\mathcal{G}_X$ of directed acyclic graphs with hidden variables such that for any $G \in \mathcal{G}_X$, for any causal Bayes net $\cP$ on $G$, and for any intervention $x$ to $X$, $P_x$ is identifiable from $P$. Thus, for all our work we assume that $G \in \mathcal{G}_X$, because otherwise, $P_x$ is not identifiable, even with an infinite number of observations.

We design sample and time efficient algorithms for the above-mentioned estimation problems. 
Our starting point is the work of Tian and Pearl~\cite{TP02b}.   Tian and Pearl~\cite{TP02b} (as well as other related work on identifiability)  assumes, in addition to the underlying graph being in  $\mathcal{G}_X$, that the distribution $P$ is {\em positive}, meaning that $P(\vec{v}) > 0$ for all assignments $\vec{v}$ to $\vec{V}$. We show that under reasonable assumptions about the structure of $G$, we only need to assume {\em strong positivity} for the marginal of $P$ over a bounded number of variables to design our algorithms.  
We extend our techniques to the problem of efficiently estimating the marginal interventional distributions over a subset of observable variables. Finally we establish a lower bound for the sample complexity showing that our sample complexity has near optimal dependence on the parameters of interest. We discuss our results in detail next. 

\ignore{
We aim to derive a minimal set of additional assumptions such that the above estimation problems admit both sample and time efficient algorithms. \cite{TP02b} as well as later related works all assume that the distribution $P$ is {\em positive}, meaning that $P(\vec{v}) > 0$ for all assignments $\vec{v}$ to $\vec{V}$. This is clearly a strong assumption, especially if the number of variables in $\vec{V}$ is large. We show that under reasonable assumptions about the structure of $G$, we only need to assume {\em strong positivity} for the marginal of $P$ over a bounded number of variables. Moreover, under this assumption, there exist sample and time-efficient algorithms for each of the above problems. }


\section{Our Contributions} \label{contributions}

\begin{figure}
\centering
\includegraphics{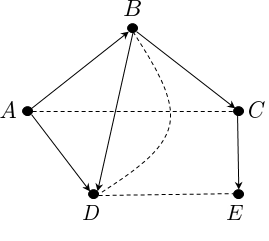}
\caption{An acyclic directed mixed graph (ADMG) where the bidirected edges are depicted as dashed. The in-degree of the graph is 2. The c-components are $\{A,C\}$ and $\{B,D,E\}$.}
\label{fig:admg}
\end{figure}

Let $\cP$ be a causal Bayes net\footnote{Formal definitions appear in \cref{sec:prelim}.} over a graph $G$, in which the set of observable variables is denoted by $\vec{V}$ and the set of hidden variables is denoted by $\vec{U}$. 
Let $n = |\vec{V}|$. There is a standard procedure in the causality literature (see \cite{TP02a}) to convert $G$ into a graph on $n$ nodes. Namely, under the {\em semi-Markovian} assumption that each hidden variable $U$ does not have any parents and affects exactly two observable variables $X_i$ and $X_j$, we remove $U$ from $G$ and put a {\em bidirected} edge between $X_i$ and $X_j$. We end up with an {\em Acyclic Directed Mixed Graph (ADMG)} $G$, having  $n$ nodes corresponding to the variables $\vec{V}$ and having edge set $E^\to \cup E^\leftrightarrow$ where $E^\to$ are the directed edges and $E^\leftrightarrow$ are the bidirected edges. \cref{fig:admg} shows an example. The {\em in-degree} of $G$ is the maximum number of directed edges coming into any node.
A {\em c-component} refers to any maximal subset of nodes/variables which is connected using only bidirected edges. Then $\vec{V}$ gets partitioned into c-components: $\vec{S}_1, \vec{S}_2,\dots,\vec{S}_\ell$. 

Let $X$ be a designated variable in $\vec{V}$. Without loss of generality, suppose $X \in \vec{S}_1$. 
Tian and Pearl~\cite{TP02b} showed that $\mathcal{G}_X$ (\ignore{defined above as }the class of ADMGs $G$ for which $P_x$ is identifiable from $P$ for any causal Bayes net $\cP$ on $G$ and any intervention $x \in \Sigma$) consists of exactly those graphs that satisfy \cref{ass:id} below (See Theorem~3 of \cite{TP02b}). 

\begin{assumption}[Identifiability with respect to $X$]\label{ass:id}
There does not exist a path of bidirected edges between $X$ and any child of $X$. Equivalently, no child of $X$ belongs to $\vec{S}_1$.
\end{assumption}

The second assumption we make is about the observational distribution $P$. For a subset of variables $\vec{S} \subseteq \vec{V}$, let $\Pa^+(\vec{S}) = \vec{S} \cup \bigcup_{V \in \vec{S}} \Pa(V)$ where $\Pa(V)$ are the observable parents of $V$ in the graph $G$.
\begin{assumption}[$\alpha$-strong positivity with respect to $X$]\label{ass:bal}
Suppose $X$ lies in the c-component $\vec{S}_1$, and let $\vec{Z} = \Pa^+(\vec{S}_1)$. For every assignment $\vec{z}$ to $\vec{Z}$, $P(\vec{Z}=\vec{z}) > \alpha$. 
\end{assumption}
So, if $|\Pa^+(\vec{S}_1)|$ is small, then \cref{ass:bal} only requires that a small set of variables take on each possible configuration with non-negligible probability. When~\cref{ass:bal} holds, we say that the causal Bayes net is {\em $\alpha$-strongly positive with respect to $X$}. More generally, if an observational distribution $P$ satisfies  $\forall \vec{s}, P(\vec{S}=\vec{s}) > \alpha$ for some $\alpha>0$ and subset $\vec{S}$ of variables we say $P$ is {\em $\alpha$-strongly positive with respect to $\vec{S}$}.

\subsection{Algorithms}
Suppose $\cP$ is an unknown causal Bayes net over a known ADMG $G$ on $n$ observable variables $\vec{V}$ that satisfies identifiablity (\cref{ass:id}) and $\alpha$-strong positivity (\cref{ass:bal}) with respect to a variable $X \in \vec{V}$ . Let $d$ denote the maximum in-degree of the graph $G$ and $k$ denote the size of its largest c-component. 

 We present an efficient algorithm for the evaluation and generation problems. 
\begin{restatable}{theorem}{learn}[Evaluation and Generation\footnote{
All our learning algorithms succeed with $1-\delta$ probability and the sample and the time complexity dependences are $O(\log {1\over \delta})$ and $O(\log^3  {1\over \delta})$ respectively for any $0<\delta<1$.  
}]\label{thm:learn}
For any intervention $x$ to $X$ and parameter $\eps \in (0,1)$, there is an algorithm that takes $m = \tilde{O}\left(\frac{|\Sigma|^{2kd}n}{\alpha^k\epsilon^{2}}\right)$ samples from $P$, and in $\tilde{O}\left(\frac{|\Sigma|^{4kd}n^2}{\alpha^k\epsilon^2}\right)$ time, learns a distribution $\hat{P}$ satisfying $\dtv(P_x,\hat{P})\le \epsilon$ such that
\begin{itemize}
\item{Evaluation:} Given an assignment $\vec{w}$ to $\vec{V}\setminus \{X\}$ computing $\hat{P}(\vec{w})$ takes $O(n|\Sigma|(kd+k))$ time 
\item{Generation:} Obtaining an independent sample from $\hat{P}$ takes $O(n|\Sigma|(kd+k))$ time .
\end{itemize}
\end{restatable}

We now discuss the problem of estimating $P_x|_{\vec{F}}$, i.e., the marginal interventional distribution upon intervention $x$ to $X$ over a subset of the observables $\vec{F} \subseteq \vec{V}$.  We show finite sample bounds for estimating $P_x|_{\vec{F}}$ when the causal Bayes net satisfies \cref{ass:id} and \cref{ass:bal}, thus obtaining quantitative counterparts to the results shown in \cite{TP02b} (See Theorem~4 of \cite{TP02b}).  We use $f$ to denote the cardinality of $\vec{F}$. 

A generator for $P_x$ obviously also gives a generator for the marginal of $P_x$ on any subset $\vec{F}$.
We observe that given a generator, we can also learn an approximate evaluator for the marginal of $P_x$ on $\vec{F} $ sample-efficiently. This is because using $O(|\Sigma|^{f}/\eps^2)$ samples of $\hat{P}_x$, we can learn an explicit description of $\hat{P}_x|_{\vec{F}}$ upto total variation distance $\eps$ with probability at least $9/10$, by simply using the empirical estimator. Since $\hat{P}_x$ is itself $\eps$-close to $P_x$ in total variation distance, we get an algorithm that with constant probability, returns an evaluator for a distribution that is $2\eps$-close to $P_x|_{\vec{F}}$. Summarizing:

\begin{corollary} \label{cor:learn3}
For any subset $\vec{F} \subseteq \vec{V}$ with $|\vec{F}| = f$, intervention $x$ to $X$ and parameter $\eps \in (0,1)$, there is an algorithm that takes $m = \tilde{O}\left(\frac{|\Sigma|^{2kd}n}{\alpha^k \epsilon^2}\right)$ samples from $P$ and in $O(mn|\Sigma|^{2kd})$ time returns an evaluator for a distribution $\hat{P}_x|_{\vec{F}}$ on $\vec{F}$ such that $\dtv(P_x|_{\vec{F}}, \hat{P}_x|_{\vec{F}})\leq \eps$. 
\end{corollary}

Note that the time complexity of the above algorithm is exponential in $f$ as we need to take exponential in $f$ many samples from the generator.  To handle problems that arise in practice for $\vec{F}$'s of small cardinality, it is of interest to develop sample and time efficient algorithms for estimating $P_x|_{\vec{F}}$.  In such cases the approach discussed above is superfluous, as the sample complexity depends linearly on $n$, the total number of variables in the model, which could be potentially large.  We show that in such cases where $f$ is extremely small we can perform {\em efficient} estimation with {\em small sample size}.  A more detailed discussion of our analysis on evaluating marginals which includes the algorithms and proofs can be found in \cref{identifying-marginals}.  Precisely, we show the following theorem:
\begin{restatable}{theorem}{learnmarg} \label{thm:learn3}
For any subset $\vec{F} \subseteq \vec{V}$ with $|\vec{F}| = f$, intervention $x$ to $X$ and parameter $\eps \in (0,1)$, there is an algorithm that takes $m = \tilde{O}\left(\frac{|\Sigma|^{2(f + k(d+1))^2} \ignore{(f + k(d+1))}}{\alpha^k\epsilon^2}\right)$ samples from $P$ and runs in $O(m(f + k(d+1)) |\Sigma|^{2(f + k(d+1))^2})$ time and returns an evaluator for a distribution $\tilde{P}_{\vec{F}}$ on $\vec{F}$ such that $\dtv(P_x|_{\vec{F}}, \tilde{P}_{\vec{F}})\leq \eps$. 
\end{restatable}

\subsection{Lower Bounds}
We next address the question of whether the sample complexity of our algorithms has the right dependence on the parameters of the causal Bayes net as well as on $\alpha$. We also explore whether \cref{ass:bal} can be weakened. Since in this section, our focus is on the sample complexity instead of time complexity, we do not distinguish between evaluation and generation.

\begin{figure*}[t]
\centering
\begin{subfigure}[t]{0.4\textwidth}
\begin{tikzpicture}
\node[text width=0.3cm] at (0,0) {};
\node[text width=0.3cm] at (3,0) {$X$};
\node[text width=0.3cm] at (5,0) {$Y$};
\draw [arrow] (3.2,0) -- (4.8,0);
\end{tikzpicture}

\caption{Lower bound for when $X$ is a source}
\label{fig:lb1}
\end{subfigure}
~
\begin{subfigure}[t]{0.4\textwidth}
\centering
\begin{tikzpicture}
\node[text width=0.3cm] at (0,0) {};
\node[text width=0.3cm] at (3,0) {$X$};
\node[text width=0.3cm] at (5,0) {$Y$};
\node[text width=0.3cm] at (4,-1) {$Z$};

\draw [arrow] (3.2,0) -- (4.8,0);
\draw [arrow] (4,-0.8) -- (3.2,0);
\draw [arrow] (4,-0.8) -- (4.8,0);

\end{tikzpicture}

\caption{Lower bound for when $X$ has a parent}
\label{fig:lb2}
\end{subfigure}
\caption{}
\end{figure*}

To get some intuition, consider the simple causal Bayes net depicted in \cref{fig:lb1}. Here, $X$ does not have any parents and $X$ is not confounded with any variable. $Y$ is a child of $X$, and suppose $X$ and $Y$ are boolean variables, where $P(X=0)=\alpha$ for some small $\alpha$. Now, to estimate the interventional probability $P_{X=0}(Y=0) =P(Y=0\mid X=0)$ to within $\pm \eps$, it is well-known that $\Omega(\eps^{-2})$ samples $(X,Y)$ with $X=0$ are needed. Since $X=0$ occurs with probability $\alpha$, an $\Omega(\alpha^{-1}\eps^{-2})$ lower bound on the sample complexity follows.

However, from this example, it's not clear that we need to enforce strong positivity on the parents of $X$ or the c-component containing $X$, since both are trivial. Also, the sample complexity has no dependence on $n$ and $d$. The following theorem addresses these issues. 
\begin{restatable}{theorem}{lbmain}\label{thm:lbmain}
Fix integers $d,k \geq 1$ and a set $\Sigma$ of size $\geq 2$. 
For all sufficiently large $n$, there exists an ADMG $G$ with $n$ nodes and in-degree $d$ so that the following hold. $G$ contains a node $X$ such that $|\Pa(X)|=d$ and $|\vec{S}_1|=k$ (where $\vec{S}_1$ is the c-component containing $X$). For any $Z \in \Pa(X) \cup \vec{S}_1$, there exists a causal Bayes net $\cP$ on $G$ over $\Sigma$-valued variables such that:
\begin{enumerate*}
\item[(i)] For the observational distribution $P$, the marginal $P|_{(\Pa(X) \cup \vec{S}_1)\setminus \{Z\}}$ is uniform but the marginal $P|_{\Pa(X) \cup \vec{S}_1}$ has mass at most $\alpha$ at some assignment.
\item[(ii)]  There exists an intervention $x$ on $X$ such that learning the distribution $P_x$ upto $\dtv$-distance $\eps$ with probability $9/10$ requires $\Omega(n |\Sigma|^d/\alpha \eps^2)$ samples from $P$.
\end{enumerate*}
\end{restatable}
So, $P$ must have a guarantee that its marginal on $\Pa(X) \cup \vec{S}_1$ has mass $>\alpha$ at all points in order for an algorithm to learn $P_x$ using $O(n |\Sigma|^d/\alpha \eps^2)$ samples. For comparison, our algorithms in \cref{thm:learn} assume strong positivity for $\Pa^+(\vec{S}_1)$ and achieve sample complexity $O(n |\Sigma|^{5kd}/\alpha^k\eps^2)$. For small values of $k$ and $d$, the upper and lower bounds are close. It remains an open question to fully close the gap.

To hint towards the proof of \cref{thm:lbmain}, we sketch the argument when $Z$ is a parent of $X$ and $n=3$. \cref{fig:lb2} shows a graph where $X$ has one parent $Z$ and no hidden variables. Both $X$ and $Z$ are parents of $Y$, and all three are binary variables. Consider two causal models $\cP$ and $\cQ$. For both $P$ and $Q$, $Z$ is uniform over $\{0,1\}$ and $X\neq Z$ with probability $\alpha$. Now, suppose $P(Y=1 \mid X \neq Z)= 1/2 + \eps$ and $P(Y=1 \mid X = Z)=1/2$, while $Q(Y=1 \mid X \neq Z)= 1/2 - \eps$ and $Q(Y=1 \mid X = Z)=1/2$. Note that $P_{X=1}(Y=1)= (1+\eps)/2$ while $Q_{X=1}(Y=1)=( 1-\eps)/2 $, so that the interventional distributions are $\epsilon$-far from each other. On the other hand, it can be shown using Fano's inequality that any algorithm needs to observe $\Omega(\alpha^{-1}\eps^{-2})$ samples to distinguish $P$ and $Q$.

\subsection{Previous Work}

Identification of  causal effects  from the observational distribution
has been studied extensively in the literature.  Here we discuss some of the relevant literature in the non-parametric setting. When there are no unobservable variables (and hence the associated ADMG is a DAG), it is always possible to identify any given intervention  from the observational distribution~\cite{pearl00,robins86,SGS00}.  However, when there are unobservable variables causal effect identifiability in ADMGs is not always possible.  A  series of important works focused on establishing graphical criterions for identifiability of interventional distributions from the observational distribution~\cite{TP02b,SGS00,GP95,Halpern00,KM99,PR95}. This led to a complete algorithm\footnote{Complete algorithms output the desired causal effect whenever possible or output fail along with a proof of unidentifiability -- thus characterizing a necessary and sufficient graphical condition for identifiability.}, first by Tian and Pearl for the identifiability of atomic interventions \cite{TP02b} (this work is the most relevant for the present work), and then by Shpitser and Pearl (algorithm {\rm ID}) for the identifiability of any given intervention from the observational distribution~\cite{SP06} (see also \cite{HV08}). Researchers have also investigated implementation aspects of the identification algorithms.  In particular, an implementation of the algorithm {\rm ID} has been carried out in the R package {\tt causaleffect} in \cite{TK17a}.  This work was followed by a sequence of works \cite{TK17b,TK18} where the authors simplify {\rm ID} and obtain a succinct representation of the target causal effect by removing unnecessary variables from the expression.
Other software packages related to causal identifiability are also publicly available \cite{jintian, akelleh, dowhy}.  

Researchers have also investigated non-parametric causal effect identification from observations on structures other than ADMGs. Some recent results in this direction include work reported in \cite{JZB19a} (and \cite{JZB19b}) where complete algorithms have been established for causal effect identifiability (and conditional causal effect identifiability) with respect to {\em Markov equivalent class diagrams}, a more general class of causal graphs.  {\em Maximally oriented partially directed acyclic graphs} (MPDAGs) is yet another generalization of DAGs with no hidden variables. Very recently complete algorithms for causal identification with respect to MPDAGs have been established \cite{Perkovic19}.  Complete algorithms are also known for {\em dynamic causal networks}, a causal analogue for dynamic Bayesian networks that evolve over time~\cite{BAG16}.  {\em Causal chain graphs} (CEGs, which are similar to ADMGs) are yet another class of graphs  for which identifiability of interventions has been investigated and conditions (similar to Pearl's back-door criterion) have been established~\cite{TSR10, Thwaites13}.

In a different line of work reported in \cite{SS16}, the authors  introduce the notion of {\em stability of causal identification}: a notion capturing the sensitivity of causal effects to small perturbations in the input. They show that the causal identification function is numerically unstable for the {\rm ID} algorithm~\cite{SP06}. They also show that, in contrast  for atomic interventions (i.e., when $X$ is singleton) the identification algorithm of Tian and Pearl~\cite{TP02b} is not too sensitive to changes in the input whenever~\cref{ass:id} of \cite{TP02b} is true. \ignore{ This work paves way to the interesting followup question: whether are there efficient algorithms for estimating an unknown causal effect from finite samples -- under the identifiability condition of \cite{SP06}?  If so, under what conditions is it possible to estimate the desired causal effect?}  

Although most of the work on non-parametric causal identification mentioned above assume the causal graph is known, the problem of inferring the underlying causal graph has also been studied in  various contexts. Some papers reporting the work along this line include
~\cite{HEJ15,HB13,asysu19,yku18,kocaoglu2019characterization}.
Causal effect identification is a fundamental topic with a wide range of practical applications. In particular it has found applications in a range of applied areas including recommendation systems \cite{SHW15}, computational sciences \cite{spirtes10}, social and behavioral sciences \cite{Sobel00}, econometrics \cite{HV07,Matzkin93,Lewbel19}, and epidemiology \cite{hernan-book}.

An important observation we note is that all existing works on non-parametric causal identifiability research assume infinite sample access to the observational distribution.  To the best of our knowledge, the present work is the first that
establishes sample and time complexity bounds on non-parametric causal effect identifiability. In this respect, the closest
related work is \cite{ABDK18} which looked at the problem of goodness-of-fit testing of causal models in a 
non-parametric setting; however, they assumed access to experimental data, not just observational data. Jung et al.~\cite{DBLP:conf/aaai/Jung0B20} gave a weighting-based estimator for expected causal effects under certain graphical conditions.
\section{Preliminaries}\label{sec:prelim}

\paragraph{Notation.} 
We use capital (bold capital) letters to denote variables (sets of variables), e.g., $A$ is a variable and $\mv{B}$ is a set of variables. We use small (bold small) letters to denote values taken by the corresponding variables (sets of variables), e.g., $a$ is the value of $A$ and $\mv{b}$ is the value of the set of variables $\mv{B}$. For a vector $\vec{v}$ and a subset of coordinates $\vec{S}$, we use the notation $\vec{v}_{\vec{S}}$ to denote the restriction of $\vec{v}$ to the coordinates in $\vec{S}$ and $v_i$ to denote the $i$-th coordinate of $\vec{v}$. For two sets of variables $\mv{A}$ and $\mv{B}$ and assignments of values  $\mv{a}$ to $\mv{A}$ and $\mv{b}$ to $\mv{B}$, $\mv{a}\circ\mv{b}$ (also $\mv{a},\mv{b}$) denotes the assignment to $\mv{A}\cup \mv{B}$ in the natural way. 

The variables in this paper take values in a finite set $\Sigma$. We use the {\em total variation distance} to measure the distances between distributions. For two distributions  $P$ and $Q$ over the same finite sample space $[D]$, their total variation distance is denoted by  $\dtv(P,Q)$ and is given by $\dtv(P,Q):=\frac12 \sum_{i\in [D]} |P(i)-Q(i)|.$ The KL distance between them is defined as $\sum_i P(i) \ln {P(i)\over Q(i)}$. Pinsker's inequality says $\dtv(P,Q) \le \sqrt{2 \dkl(P,Q)}$.

\paragraph{Bayesian Networks.}
Bayesian networks are popular probabilistic graphical models for describing high-dimensional distributions.

\begin{definition}
A {\em Bayesian Network} $P$ is a distribution that can be specified by a tuple $\langle \mv{V}, G, \{\Pr[V_i \mid \pa(V_i)]: V_i \in \mv{V}, \pa(V_i) \in \Sigma^{|\Pa(V_i)|}\} \rangle$ where: (i) $\mv{V}=(V_1,\dots,V_n)$ is a set of variables over alphabet $\Sigma$, (ii) $G$ is a directed acyclic graph with $n$ nodes corresponding to the elements of $\mv{V}$, and (iii) $\Pr[V_i \mid \pa(V_i)]$ is the conditional distribution of variable $V_i$ given that its parents $\Pa(V_i)$ in $G$ take the values $\pa(V_i)$. 

The Bayesian Network 
$P =  \langle \mv{V}, G, \{\Pr[V_i\mid \pa(V_i)]\} \rangle$ 
defines a probability distribution over $\Sigma^{|\mv{V}|}$, as follows. For all $\mv{v} \in \Sigma^{|\mv{V}|}$,
$$P(\mv{v}) = \prod_{V_i \in \mv{V}} \Pr[v_i  \mid  \Pa(V_i)=\mv{v}_{\Pa(V_i)}].$$
In this distribution, each variable $V_i$ is independent of its non-descendants given its parents in $G$. 
\end{definition}
\ignore{
Conditional independence relations in graphical models are captured by the following definitions.

\begin{definition}
Given a DAG $G$, a (not necessarily directed) path $p$ in $G$ is said to be {\em blocked} by a set of nodes $\mathbf{Z}$, if (i) $p$ contains a chain node $B$ ($A \rightarrow B \rightarrow C$) or a fork node $B$ ($A \leftarrow B \rightarrow C$) such that $B \in \mathbf{Z}$ (or) (ii) $p$ contains a collider node $B$ ($A \rightarrow B \leftarrow C$) such that $B \notin \mathbf{Z}$ and no descendant of $B$ is in $\mathbf{Z}$. 
\end{definition}

\begin{definition}[d-separation]
For a given DAG $G$ on $\mv{V}$, two disjoint sets of vertices $\mathbf{X,Y} \subseteq \mv{V}$ are said to be {\em d-separated} by $\mathbf{Z}$ in $G$, if every (not necessarily directed) path in $G$ between $\mathbf{X}$ and $\mathbf{Y}$ is blocked by $\mathbf{Z}$.
\end{definition}

\begin{lemma}[Graphical criterion for independence] 
For a given $\mathcal{N} = \langle \mv{V}, G, \{\Pr[V_i  \mid  \pa(V_i)]\}\rangle$ and $\mv{X,Y,Z} \subset \mv{V}$, if $\mathbf{X}$ and $\mathbf{Y}$ are {d-separated} by $\mathbf{Z}$ in $G$, then $\mathbf{X}$ is {\em independent} of $\mathbf{Y}$ given $\mathbf{Z}$ in $P_{\mathcal{N}}$, denoted by $[\mathbf{X} \indep \mathbf{Y}  \mid  \mathbf{Z}]$ in $P_{\mathcal{N}}$.
\end{lemma}             
}
\paragraph{Causality.}
We describe Pearl's notion of causality from~\cite{Pearl95}. Central to his formalism is the notion of an {\em intervention}. Given an observable variable set $\mv{V}$ and a subset $\mv{X} \subset \mv{V}$, an intervention $\cdo(\mv{x})$ is the process of fixing the set of variables $\mv{X}$ to the values $\mv{x}$. The {\em interventional distribution} $P_{\vec{x}}$ is the distribution on $\mv{V}$ after setting $\mv{X}$ to $\mv{x}$. Formally:

\ignore{
Another important component of Pearl's formalism is that some variables may be hidden (latent). The hidden variables can neither be observed nor be intervened. 
Let $\mv{V}$ and $\mv{U}$ denote the observable and hidden variables respectively. Given a directed acyclic graph $H$ on $\mv{V \cup U}$ and a subset $\mv{X} \subseteq (\mv{V \cup U})$, we use $\bm{\Pi}_H(\mv{X}), \Pa_H(\mv{X})$, $\An_H(\mv{X})$, and $\De_H(\mv{X})$  to denote the set of all parents, observable parents, observable ancestors and observable descendants respectively of $\mv{X}$, excluding $\mv{X}$, in $H$. When the graph $H$ is clear, we may omit the subscript. As usual, small letters, $\bm{\pi}(\mv{X}),$ $\pa(\mv{X})$, $\an(\mv{X})$ and $\de(\mv{X})$  are used to denote their corresponding values. 
}
\begin{definition}[Causal Bayes Net] \label{def:causal Bayesnet}
A {\em causal Bayes net} $\cP$  is a collection of interventional distributions that can be defined in terms of a tuple $\langle \mv{V}, \mv{U}, G,$ $\{\Pr[V_i \mid \bm{\pi}(V_i)] : V_i \in \mv{V}, \bm{\pi}(V_i) \in \Sigma^{|\bm{\Pi}(V_i)|}\}, \{\Pr[\vec{U}]\}\rangle$, where (i) $\mv{V}=(V_1,\dots, V_n)$ and $\mv{U}$ are the tuples of observable and hidden variables respectively, (ii) $G$ is a directed acyclic graph on $\mv{V} \cup \mv{U}$, (iii) $\Pr[V_i \mid \bm{\pi}(V_i)]$ is the conditional probability distributions of $V_i \in \vec{V}$ given that its parents $\bm{\Pi}(V_i) \in \vec{V} \cup \vec{U}$ take the values $\bm{\pi}(V_i)$, and (iv) $\Pr[\vec{U}]$ is the distribution of the hidden variables $\vec{U}$.  $G$ is said to be the {\em causal graph} corresponding to $\cP$.

Such a causal Bayes net $\cP$ 
defines a unique interventional distribution $P_{\vec{x}}$ for every subset $\mv{X} \subseteq \mv{V}$ (including $\mv{X} = \emptyset$) and assignment $\mv{x} \in \Sigma^{|\mv{X}|}$, as follows. For all $\mv{v} \in \Sigma^{|\mv{V}|}$:
$$P_{\vec{x}}(\mv{v}) = 
\begin{cases}
\sum_{\mv{u}} \prod_{{V_i} \in \mv{V}\setminus \mv{X}} \Pr[v_i \mid \bm{\Pi}(V_i)=\vec{v}_{\bm{\Pi}(V_i)}] \cdot \Pr[\vec{u}] &
\text{if }\mv{v} \text{ is consistent with }\mv{x}\\
0 &\text{ otherwise.}
\end{cases}
$$
We use $P$ to denote the observational distribution ($X = \emptyset$). For a subset $\vec{F} \subseteq \vec{V}$, $P_{\vec{x}}|_{\vec{F}}$ denotes the marginal of $P_{\vec{x}}$ on $\vec{F}$. For an assignment $\vec{f}$ to $\vec{F}$, we also use the notation $P_{\vec{x}}(\vec{f})$ as shorthand for the probability mass of $P_{\vec{x}}|_{\vec{F}}$ at $\vec{f}$.
\end{definition}
 As mentioned in the introduction, we often consider a causal graph $G$ as an ADMG by implicitly representing hidden variables using bidirected edges. In an ADMG, we imagine that there is a hidden variable subdividing each such bidirected edge that is a parent of the two endpoints of the edge. Thus, the edge set of an ADMG is the union of the directed edges $E^{\to}$ and the bidirected edges $E^{\leftrightarrow}$.  Given such an ADMG $G$, for any $\vec{S} \subseteq \vec{V}$, $\vec{\overline{S}}$ denotes the complement set $\vec{V \setminus S}$, $\Pa(\vec{S})$ denotes the parents of $\vec{S}$ according to the directed edges of $G$, i.e., $\Pa(\vec{S}) = \cup_{X \in \vec{S}}\{Y \in \vec{V}: (Y,X) \in E^{\to}\}$.  We also define: $\Pa^+(\vec{S}) = \Pa(\vec{S}) \cup \vec{S}$ and $\Pa^-(\vec{S}) = \Pa(\vec{S}) \setminus \vec{S}$. 
 The bidirected edges are used to define c-components:

\begin{definition}[c-component] \label{defn:c-component}
For a given ADMG $G$, $\mv{S} \subseteq \mv{V}$ is a {\em c-component} of $G$, if $\mv{S}$ is a maximal set such that between any two vertices of $\mv{S}$, there exists a path that uses only the bidirected edges $E^{\leftrightarrow}$.
\end{definition}

Since a c-component forms an equivalence relation, the set of all c-components forms a partition of $\mv{V}$, the observable vertices of $G$.  Let $\mv{S}_1 \cup \mv{S}_2\cup \cdots \cup \mv{S}_\ell$ denote the partition of $\mv{V}$ into the c-components of $G$. 

\begin{definition}
For a subset $\vec{S} \subseteq \vec{V}$, the {\em Q-factor for $\vec{S}$} is defined as the following function over $\Sigma^{|\vec{V}|}$:
$$Q_{\vec{S}}(\vec{v}) = P_{\vec{v}_{\overline{\vec{S}}}}(\vec{v}_{\vec{S}}).$$ 
Clearly, for every $\vec{v}_{\overline{\vec{s}}}$, $Q_{\vec{S}}$ is a distribution over $\Sigma^{|\vec{S}|}$.
\end{definition}

For $\mv{Y} \subseteq \mv{V}$, the induced subgraph $G[\mv{Y}]$ is the subgraph obtained by removing the vertices $\mv{V \setminus Y}$ and their corresponding edges from $G$.  \ignore{We use the notation $C(\mv{Y}) = \{\mv{S}_1,\mv{S}_2,\ldots,\mv{S}_k\}$ to denote the set of all c-components of $G[\mv{Y}]$, that is each $\mv{S}_i \subseteq \mv{Y}$ is a c-component of $G[\mv{Y}]$. }

The following lemma is used heavily in this work.
\begin{lemma}[Corollary 1 of \cite{tian-thesis}]\label{lem:imp}
Let $\cP$ be a causal Bayes net on $G=(\vec{V},E^\to \cup E^\leftrightarrow)$. Let  $\vec{S}_1, \dots, \vec{S}_\ell$ be the c-components of $G$. Then for any $\vec{v}$ we have:
\begin{itemize}
\item[(i)] $P(\vec{v}) = \prod_{i=1}^\ell Q_{\vec{S}_i}(\vec{v})$.
\item[(ii)] Let $V_1, V_2,\cdots, V_n$ be  a topological order over $\vec{V}$ with respect to the directed edges. Then,  for any $j \in [\ell]$,  $Q_{\vec{S}_j}(\vec{v})$  is computable from $P(\vec{v})$ and is given by:
$$Q_{\vec{S}_j}(\vec{v}) = \prod_{i: V_i \in \vec{S}_j} P(v_i \mid v_1, \dots, v_{i-1}).$$
\item[(iii)] Furthermore, each factor $P(v_i \mid v_1, \dots, v_{i-1})$ can be expressed as:
$$P(v_i \mid v_1, \dots, v_{i-1}) = P(v_i \mid \vec{v}_{\Pa^+(\vec{T}_i) \cap [i-1]})$$
where $\vec{T}_i$ is the c-component of $G[V_1, \dots, V_i]$ that contains $V_i$.  
\end{itemize}
\end{lemma}
Note that \cref{lem:imp} implies that each $Q_{\vec{S}_j}(\vec{v})$ is a function of the coordinates of $\vec{v}$ corresponding to $\Pa^+(\vec{S}_j)$. The next result, due to Tian and Pearl, uses the identifiability criterion encoded in \cref{ass:id}.
\begin{theorem}[Theorem 3 of \cite{TP02b}]\label{thm:tp}
Let $\cP$ be a causal Bayes net over $G=(\vec{V},E^\to \cup E^\leftrightarrow)$ and $X\in \vec{V}$ be a variable. Let  $\vec{S}_1, \dots, \vec{S}_\ell$ be the c-components of $G$ and assume $X\in \vec{S}_1$ without loss of generality. Suppose $G$ satisfies  \cref{ass:id} (identifiability with respect to $X$).  Then for any setting $x$ to $X$ and any assignment $\vec{w}$ to $\vec{V}\setminus \{X\}$, the interventional  distribution $P_x(\vec{w})$  is given by: 
\begin{align*}
P_x(\vec{w})
&= P_{\vec{w}_{\vec{V}\setminus \vec{S}_1}}(\vec{w}_{\vec{S}_1\setminus \{X\}}) \cdot \prod_{j=2}^\ell P_{\vec{w}_{\vec{V}\setminus (\vec{S}_j\cup \{X\})}\circ x}(\vec{w}_{\vec{S}_j})\\
&= \sum_{x' \in \Sigma}Q_{\vec{S}_1}(\vec{w}\circ x') \cdot \prod_{j=2}^\ell Q_{\vec{S}_j}(\vec{w}\circ x)
\end{align*} 
\end{theorem}
\section{Efficient Estimation}
Let $\cP$ be a causal Bayes net over a causal graph $G=(\vec{V},E^\to \cup E^\leftrightarrow)$. $G$ is an ADMG with 
observable variables $V_1, \dots, V_n$. Without loss of generality, let $V_1, \dots, V_n$ be a topological order according to the directed edges of $G$. Before we proceed to our algorithms for interventional distributions, we will first present an algorithm for {\em learning the observational distribution} $P(\vec{V})$. Our approach is to then view the causal Bayes net as a regular Bayes net over observable variables and use the learning algorithm for Bayes nets. 
From \cref{lem:imp}, we can write the observational distribution $P(\vec{V})$ as:
\begin{equation}\label{eqn:obsfactor}
P(\vec{V}) = \prod_{i=1}^n P(V_i \mid \vec{Z}_i)
\end{equation}
where $\vec{Z}_i \subseteq \{V_1, \dots, V_{i-1}\}$ is the set of `effective parents' of $V_i$ of size at most $kd+k$. Here $k$ is the maximum c-component size and $d$ is the maximum in-degree. Therefore the observational distribution $P$ can also be viewed as the distribution of a (regular) Bayes net with {\em no hidden variables} but with in-degree at most $kd+k$.  The problem of properly learning a Bayes net is well-studied~\cite{Das97,Canonne}, starting from Dasgupta's early work~\cite{Das97}. In this work, we will use the following learning result described in \cite{BGMV20}.
\ignore{We use the formulation given in\footnote{The analysis in Appendix A.1 of \cite{Canonne} only works for success probability 2/3. However, it is straightforward to modify it to get the version here.}.}
\begin{theorem}[\cite{BGMV20}]\label{thm:bnlearning}
There is an algorithm that on input parameters $\eps, \delta \in (0,1)$ and samples from an unknown Bayes net $P$ over $\Sigma^n$ on a known DAG $G$ on vertex set $[n]$ and maximum in-degree $d$, takes $m=O(\log {1\over \delta}|\Sigma|^{d+1}n\log (n|\Sigma|^{d+1})/\epsilon^2)$ samples, runs in time $O(mn|\Sigma|^{d+1})$, and produces a Bayes net $\hat{P}$ on $G$ such that $\dtv(P,\hat{P})\le \epsilon$ with probability $\geq 1-\delta$.
\end{theorem}
From the above discussion we get the following corollary.

\begin{corollary}
There is an algorithm that on input parameters $\eps, \delta \in (0,1)$, and samples from the observed distribution  $P$ of an unknown causal Bayes net over $\Sigma^n$ on a known ADMG $G$ on vertex set $[n]$ with maximum in-degree $d$ and maximum c-component size $k$, takes $m=\tilde{O}(\frac{n}{\epsilon^2} |\Sigma|^{kd+k+1}\log{1 \over \delta})$ samples, runs in time $O(mn|\Sigma|^{kd+k+1})$ and outputs a Bayes net $\hat{P}$ on a DAG $G'$ such that $\dtv(P,\hat{P})\allowbreak\le \epsilon$ with probability $ \geq 1-\delta$.
\end{corollary}

\ignore{
\begin{corollary}
Let ${P}$ be an unknown causal Bayes net over a known graph $G$ on $n$ variables with maximum in-degree $d$ and maximum c-component size $k$. Then For any $\eps, \delta \in (0,1)$, the distribution on the observable set $V$ can be learnt to within $\dtv$ distance at most $\epsilon$ by taking $m=\tilde{O}(\log {1\over \delta}|\Sigma|^{kd+k+1}n/\epsilon^2)$ samples from $P$ and time $O(mn|\Sigma|^{kd+k+1})$ with probability $1-\delta$.
\end{corollary}
}

In the next subsection we design our evaluation and generation algorithms. 
\ignore{
We use the following partitioning of the variables in these subsections. 
Let the c-components of $G$ be given by $\vec{S}_1, \dots, \vec{S}_\ell$ with $X \in \vec{S}_1$.
Let $\vec{A} = \vec{S}_1\setminus \{X\}, \vec{B} = \Pa^{-}(\vec{S}_1)$ and $\vec{C} = \vec{V} \setminus (\vec{A} \cup \vec{B} \cup \{X\})$. Note that $|\vec{A}| \leq k$ and  $|\vec{B}|\leq kd$. 
See \cref{fig:ccomps}.

\begin{figure}
\centering
\includegraphics[scale=0.5]{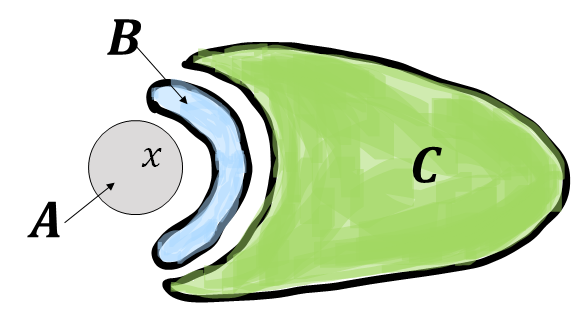}
\caption{The decomposition of the graph into $\vec{A}, \vec{B}, $ and $\vec{C}$. }
\label{fig:ccomps}
\end{figure}
}
\subsection{Evaluation and Sampling of $P_x$}
In this section we will prove~\cref{thm:learn} for learning $P_x$. Let the index of $X\in \vec{V}$ be $t$ in the topological ordering i.e. $X=V_t$. Let $\vec{S}_1$ be the c-component of $X$. According to~\cref{eqn:obsfactor}, the observational distribution $P(\vec{V})$ factorizes as a Bayes net into factors of the form $P(V_i \mid \vec{Z}_i)$ where $\vec{Z}_i$ are the effective dependants of $V_i$.  

We note that the interventional distribution $P_x$ can be represented as a marginal distribution of a different Bayes net $D_x(\vec{V})$ based on the following observation that uses~\cref{thm:tp}.  To obtain this representation of $P_x$, consider the Bayes net factorization of $P(\vec{V})$. Replace all the factors $P(V_i \mid \vec{Z}_i)$ satisfying $V_i \notin \vec{S}_1$ and $X \in \vec{Z}_i$ by $P(V_i \mid \vec{Z}_i\setminus \{X\},X=x)$. In other words, each of these factors now does not use the variable $X$ and instead uses the constant $X=x$ (which is the intervention) as parent. All the other factors, including $P(V_t \mid \vec{Z}_t)$, remain the same as in $P(\vec{V})$.  The marginal distribution of $D_x$ on $\vec{V}\setminus \{X\}$ is exactly $P_x$. This is illustrated in~\cref{fig:bnobs} and~\cref{fig:bndo}.

\begin{figure*}[t]
\centering
\begin{subfigure}[t]{0.49\textwidth}
\resizebox{\textwidth}{!}{
\begin{tikzpicture}
\draw (0,0) circle (1.5cm);
\draw (1.5,-2.5) arc (-60:64:3cm);
\draw (1.5,-2.5) arc (-30:34:5cm);
\draw (3,-2.5) arc (-45:45:4cm);
\draw (3,-2.5) arc (-90:-45.8:10cm);
\draw (3,3.15) arc (90:45.8:10cm);

\node at (-.25, -0.75)   (v1) {$\bullet$};
\node at (-0.25, 0.75)   (v2) {$\bullet$};
\node at (-0.5, 0.75)  (v1name) {$X$};

\node at (2.5, 0.75)   (v3) {$\bullet$};
\node at (2.5, -0.25)   (v4) {$\bullet$};

\node at (4.5, 1.75)   (v5) {$\bullet$};
\node at (4.5, -1.75)   (v6) {$\bullet$};
\node at (7, 1)   (v7) {$\bullet$};
\node at (4.5, 0)   (v7) {$\bullet$};

\draw [arrow] (4.5,0) -- (2.6, 0.75);
\draw [arrow] (2.5, 0.75) -- (-.19, -0.68);
\draw [arrow] (2.5, -0.25) -- (-0.2, 0.7);
\draw [arrow] (2.5, -0.25) -- (-.2, -0.75);
\draw [arrow] (-0.25, 0.75) -- (-.25, -0.7);
\draw [arrow] (-0.25, 0.75) -- (2.4, .75);
\draw [arrow] (2.5, 0.75) -- (4.45, 1.7);

\draw [arrow] (-0.25, 0.75) .. controls (2,5) .. (6.95,1.05);

\draw [arrow] (2.5, -0.25) -- (4.45, -1.7);
\draw [arrow] (4.5, -1.75) -- (6.95, 0.95);

\node at (0,-3) {$\mathbf{S}_1$};
\node at (2,-3) {$\mathbf{Pa}^{-}(\mathbf{S}_1)$};
\node at (6,-3) {$\mathbf{V}\setminus\mathbf{Pa}^{+}(\mathbf{S}_1)$};

\end{tikzpicture}
}
\caption{The Bayes net $P(\vec{V})$.}
\label{fig:bnobs}
\end{subfigure}
\begin{subfigure}[t]{0.49\textwidth}
\centering
\resizebox{\textwidth}{!}{
\begin{tikzpicture}
\draw (0,0) circle (1.5cm);
\draw (1.5,-2.5) arc (-60:64:3cm);
\draw (1.5,-2.5) arc (-30:34:5cm);
\draw (3,-2.5) arc (-45:45:4cm);
\draw (3,-2.5) arc (-90:-45.8:10cm);
\draw (3,3.15) arc (90:45.8:10cm);

\node at (-.25, -0.75)   (v1) {$\bullet$};
\node at (-0.25, 0.75)   (v2) {$\bullet$};
\node at (-0.5, 0.75)  (v1name) {$X$};

\node at (2.5, 0.75)   (v3) {$\bullet$};
\node at (2.5, -0.25)   (v4) {$\bullet$};

\node at (4.5, 1.75)   (v5) {$\bullet$};
\node at (4.5, -1.75)   (v6) {$\bullet$};
\node at (7, 1)   (v7) {$\bullet$};
\node at (4.5, 0)   (v7) {$\bullet$};
\node at (3,3.5) (v8) {$\bullet$};
\node at (3,3.75) (v8name) {$X=x$};

\draw [arrow] (3,3.5) -- (6.95,1.05);
\draw [arrow] (3,3.5) -- (2.5,0.8); 
\draw [arrow] (4.5,0) -- (2.6, 0.75);
\draw [arrow] (2.5, 0.75) -- (-.19, -0.68);
\draw [arrow] (2.5, -0.25) -- (-0.2, 0.7);
\draw [arrow] (2.5, -0.25) -- (-.2, -0.75);
\draw [arrow] (-0.25, 0.75) -- (-.25, -0.7);
\draw [arrow] (2.5, 0.75) -- (4.45, 1.7);


\draw [arrow] (2.5, -0.25) -- (4.45, -1.7);
\draw [arrow] (4.5, -1.75) -- (6.95, 0.95);

\node at (0,-3) {$\mathbf{S}_1$};
\node at (2,-3) {$\mathbf{Pa}^{-}(\mathbf{S}_1)$};
\node at (6,-3) {$\mathbf{V}\setminus\mathbf{Pa}^{+}(\mathbf{S}_1)$};

\end{tikzpicture}
}
\caption{The Bayes net $D_x(\vec{V})$. Variables in $\mathbf{V}\setminus \mathbf{S}_1$ which originally depended on the variable $X$ in $P$ now instead depend on the constant $X=x$.}
\label{fig:bndo}
\end{subfigure}
\caption{}
\end{figure*}

\ignore{
}

More formally, let $\vec{W}:=\vec{V}\setminus\{X\}$ and $\vec{w}$ be an arbitrary assignment to it and let $X=V_t$. Using~\cref{thm:tp}, $P_x$ can be factorized as follows:
\begin{equation}\label{eqn:doas}
P_x(\vec{w}) = \left(\sum_{x' \in \Sigma} \left(\prod_{V_i \in \vec{S}_1} P((\vec{w}\circ x')_{V_i}\mid (\vec{w}\circ x')_{\vec{Z}_i})\right)\right) \prod_{V_i \notin \vec{S}_1} P(\vec{w}_{V_i}\mid (\vec{w}\circ x)_{\vec{Z}_i})
\end{equation}
where $\vec{Z}_i$ is the effective parents of $V_i$ from~\cref{eqn:obsfactor}. So, we start with the factorization of~\cref{eqn:obsfactor} for the assignment $\vec{w}\circ x$, then replace all occurrences of $x$ with $x'$ in $\vec{Z}_i\cup \{V_i\}$ of $P(V_i \mid \vec{Z}_i)$ for every $V_i \in \vec{S}_1$ and taking a summation over all possible values of $x' \in \Sigma$.

In a view to learn $P_x$, we learn the Bayes net distribution:
\begin{equation}\label{eqn:dobn}
D_x(\vec{V}) = \prod_{\substack{V_i \in \vec{S}_1\vee\\X \notin \vec{Z}_i}} P(V_i\mid \vec{Z}_i) \prod_{\substack{V_i \notin \vec{S}_1\wedge\\X \in \vec{Z}_i}} P(V_i\mid \vec{Z}_i\setminus \{X\},x).
\end{equation}
So, we start with the factorization of~\cref{eqn:obsfactor} and replace all occurrences of the variable $X$ with the constant $x$ which appear in the factors outside of $\vec{S}_1$.
$D_x$ is a well-defined distribution as $\sum_{\vec{v}} D_x(\vec{v})=1$ by marginalizing out variables one after another in the reverse topological order, starting from the sink nodes. Learning $D_x$ suffices since its marginal on $\vec{V}\setminus \{X\}$ is exactly $P_x(\vec{W})$:
\begin{equation*}
P_x(\vec{W})=\sum_{x' \in \Sigma} D_x(\vec{W}\circ x').
\end{equation*}

We rewrite~\cref{eqn:dobn} as:
\begin{equation}\label{eqn:dobn2}
D_x(\vec{V}) = \prod_{V_i} D_x(V_i\mid \vec{Z}'_i)
\end{equation}
where $\vec{Z}'_i=\vec{Z}_i$ and $D_x(V_i\mid \vec{Z}'_i)= P(V_i\mid \vec{Z}'_i)$ for $V_i \in \vec{S}_1\vee X \notin \vec{Z}_i$; and $\vec{Z}'_i=\vec{Z}_i\setminus \{X\}$ and $D_x(V_i\mid \vec{Z}'_i)= P(V_i\mid \vec{Z}'_i,X=x)$ otherwise.

We use a KL local subadditivity result for Bayes nets from Canonne et al.~\cite{Canonne}. For a Bayes net $R$, a vertex $i$ and an assignment $\vec{a}$ to its parents, let $\Pi[i,\vec{a}]$ denote the event that the parents of a variable $i$ is $\vec{a}$ and let $R(i\mid \vec{a})$ denote the conditional distribution of variable $i$ when its parents are $\vec{a}$.
\begin{theorem}[\cite{Canonne}]\label{thm:bnadd}
Let $R, S$ be two Bayes nets over a common graph. Then 
\begin{equation}\label{eqn:bnadd}
\dkl(R,S)\le \sum_i \sum_{\vec{a}} R(\Pi[i,\vec{a}])\; \dkl(R(i\mid\vec{a}),S(i\mid\vec{a}))
\end{equation}
\end{theorem}
We also need the following result for learning the local distributions in $\dkl$ distance.
\begin{theorem}[\cite{pmlr-v40-Kamath15}]\label{thm:add1}
Let $D$ be an unknown distribution over $\Sigma$. Suppose we take $z$ samples from $D$ and define the add-1 empirical distribution $D'(i)=(z_i+1)/(z+|\Sigma|)$ where $z_i$ is the number of occurrences of item $i\in \Sigma$. Then $\ex{\dkl(D,D')}\le (|\Sigma|-1)/(z+1)$.
\end{theorem}

We are trying to learn $D_x$ but we have only sample access to $P$. The following lemma relates the p.m.f.s of $D_x$ and $P$ which we use later.
\begin{lemma}\label{lem:ineqdo}
Let $\vec{w}$ be an assignment to $\vec{V}\setminus \{X\}$ and let $x'$ and $x$ be two assignments to $X$. Suppose $P$ be $\alpha$-strongly positive w.r.t. $\Pa^{+}(\vec{S}_1)$. Then the following holds:
\begin{enumerate}
\item $P(\vec{w}\circ x) \ge \alpha^k  D_x(\vec{w}\circ {x'}) $
\item $P(\vec{w}\circ x) \ge {\alpha^k\over |\Sigma|}  D_x(\vec{w}) $
\item $P(\vec{w}) \ge {\alpha^k\over |\Sigma|}  D_x(\vec{w})$
\end{enumerate}
\end{lemma}

\begin{proof}
Let $\vec{v}=\vec{w}\circ x$ and $\vec{v'}=\vec{w}\circ x'$ .
\begin{align*}
{P(\vec{w}\circ x) \over  D_x(\vec{w}\circ {x'})} 
&={\prod_{i} P(v_i\mid \vec{v}_{\vec{Z}_i})  \over \prod_{V_i \in \vec{S}_1} P(v_i'\mid \vec{v}_{\vec{Z}_i}') \prod_{V_i \notin \vec{S}_1} P(v_i\mid \vec{v}_{\vec{Z}_i}) }\\
&= {\prod_{V_i \in \vec{S}_1} P(v_i\mid \vec{v}_{\vec{Z}_i})  \over \prod_{V_i \in \vec{S}_1} P(v_i'\mid \vec{v}_{\vec{Z}_i}') }\\
&\ge {\prod_{V_i \in \vec{S}_1} P(v_i\mid \vec{v}_{\vec{Z}_i})}\\
&\ge {\prod_{V_i \in \vec{S}_1} P(v_i, \vec{v}_{\vec{Z}_i})}\\
&\ge \alpha^k
\end{align*}
The first line uses~\cref{eqn:obsfactor} and~\cref{eqn:dobn}. The fourth line follows from $P(A\mid B)\ge P(A\mid B)P(B) = P(AB)$ for any two events $A$ and $B$. The last line follows since for $V_i\in \vec{S}_1$, $\{V_i\} \cup \vec{Z}_i \subseteq \Pa^{+}(\vec{S}_1)$ and $P$ is $\alpha$-strongly positive w.r.t. the later. 

Part 2. follows by marginalization of Part 1. over all possible $x' \in \Sigma$. Part 3. trivially follows from Part 2..
\end{proof}

Finally we show the following lemma for learning $D_x$ as as Bayes net according to the factorization of~\cref{eqn:dobn2}.

\begin{figure}
\begin{algorithm}[H]
\SetAlgoNoLine
\SetKwInOut{Input}{Input}
\SetKwInOut{Output}{Output}
\Input{Samples from $P$, parameters $m,t$}
\Output{A Bayes net $\hat{D}_x$ according to the factorization of~\cref{eqn:dobn2}}
Get $m$ samples from $P$\;
\For{every vertex $V_i\in \vec{S}_1$}{
\For{every fixing $\vec{Z}_i=\vec{a}$, where $\vec{Z}_i$ are the effective parents of $V_i$}{
$\hat{D}_x(V_i \mid \vec{Z}'_i=\vec{a}) \gets$ the add-1 empirical distribution (see~\cref{thm:add1}) at node $i$ in the subset of samples where $\vec{Z}_i=\vec{a}$\;
}
}
\For{every vertex $V_i\in \vec{V}\setminus \vec{S}_1$}{
\For{every fixing $\vec{Z}_i \setminus \{X\}=\vec{a}$, where $\vec{Z}_i$ are the effective parents of $V_i$}{
\eIf{$X \in \vec{Z}_i$}{
$N_{i,\vec{a}} \gets$ the number of samples with $\vec{Z}_i \setminus \{X\}=\vec{a}$ and $X=x$\;
\eIf{$N_{i,\vec{a}} \ge t$}{
$\hat{D}_x(V_i \mid \vec{Z}_i\setminus\{X\}=\vec{a}) \gets$ the add-1 empirical distribution at node $i$ in the subset of samples where $\vec{Z}_i \setminus X=\vec{a}$ and $X=x$\;
}{
$\hat{D}_x(V_i \mid \vec{Z}_i\setminus \{X\}=\vec{a}) \gets$ the uniform distribution over $\Sigma$\;
} 
}
{
$N_{i,\vec{a}} \gets$ the number of samples with $\vec{Z}_i =\vec{a}$\;
\eIf{$N_{i,\vec{a}} \ge t$}{
$\hat{D}_x(V_i \mid \vec{Z}_i=\vec{a}) \gets$ the add-1 empirical distribution at node $i$ in the subset of samples where $\vec{Z}_i=\vec{a}$\;
}{
$\hat{D}_x(V_i \mid \vec{Z}_i=\vec{a}) \gets$ the uniform distribution over $\Sigma$\;
} 
}
}
}
\caption{Learning $D_{x}$}
\label{algo:bn}
\end{algorithm}
\end{figure}

\begin{lemma}\label{lem:lrnpx}
Let $D_x(\vec{V})$ be the Bayes net as defined in~\cref{eqn:dobn2}. Then Algorithm~\ref{algo:bn} takes $\tilde{O}\left({n|\Sigma|^{2kd}\over\alpha^k\epsilon^2}\right)$ samples and $\tilde{O}\left({n^2|\Sigma|^{4kd} \over\alpha^k\epsilon^2}\right)$ time and returns a Bayes net $\hat{D}_x(\vec{V})$ such that $\dtv(D_x,\hat{D}_x)\le \epsilon$ with probability at least 3/4.
\end{lemma}

\begin{proof}
We run Algorithm~\ref{algo:bn} with the parameters $m=20n|\Sigma|^{kd+k+2}\allowbreak \log (n|\Sigma|^{kd+k})/(\alpha^k\epsilon^2)$ and $t=10\log (n|\Sigma|^{kd+k})$ to learn $D_x$ as $\hat{D}_x$. We rewrite~\cref{eqn:bnadd} for the distributions $D_x$ and $\hat{D}_x$ as follows :
\begin{align}\label{eqn:bnqxadd}
\dkl(D_x,\hat{D}_x)\le \sum_i \sum_{\vec{a}} D_x(\vec{Z}'_i=\vec{a})\; \dkl(D_x(V_i \mid \vec{Z}'_i=\vec{a}),\hat{D}_x(V_i \mid \vec{Z}'_i=\vec{a})) 
\end{align}
Our strategy is to learn $D_x(V_i \mid \vec{Z}'_i=\vec{a})$ by conditional sampling from either $P(\vec{V}\setminus\vec{Z}'_i\mid\vec{Z}'_i=\vec{a})$ or $P(\vec{V}\setminus(\vec{Z}'_i\cup\{X\})\mid\vec{Z}'_i=\vec{a},X=x)$ as appropriate.

First consider the summands of~\cref{eqn:bnqxadd} where $V_i \in \vec{S}_1$. In this case $D_x(V_i\mid \vec{Z}'_i)= P(V_i\mid \vec{Z}_i)$ and $\vec{Z}'_i=\vec{Z}_i \subseteq \Pa^{+}(\vec{S}_1)$ where $P$ is $\alpha$-strongly positive w.r.t. the later set. Hence at least $m\alpha/2$ samples turn up with $\vec{Z}_i=\vec{a}$ from Chernoff and union bounds except with 1/40 probability for a large enough $n$. Conditioned on this,~\cref{thm:add1} gives $\ex{\dkl(D_x(V_i \mid \vec{Z}'_i=\vec{a}),\hat{D}_x(V_i \mid \vec{Z}'_i=\vec{a}))} \le 2(|\Sigma|-1)/(m\alpha)\le \epsilon^2/(10n|\Sigma|^{kd+k})$ where $\hat{D}_x$ is the add-1 estimator on the conditional samples with $\vec{Z}_i=\vec{a}$. Since $D_x(\vec{Z}'_i=\vec{a}) \le 1$ each summand is also upper-bounded by $ \epsilon^2/(10n|\Sigma|^{kd+k})$.

Next we consider the summands $(i,\vec{a})$ with $V_i \notin \vec{S}_1$. For these summands, if $X\notin \vec{Z}_i$, we have $\vec{Z}'_i=\vec{Z}_i $, $D_x(V_i\mid \vec{Z}'_i)= P(V_i\mid \vec{Z}_i)$ and $P(\vec{Z}'_i=\vec{a}) \ge \alpha^k D_x(\vec{Z}'_i=\vec{a})/|\Sigma|$, the last inequality by marginalization of~\cref{lem:ineqdo} Part 3. over $\vec{V}\setminus(\vec{Z}'_i\cup\{X\})$.
If $X\in \vec{Z}_i$, we have $\vec{Z}'_i=\vec{Z}_i \setminus \{X\}$, $D_x(V_i\mid \vec{Z}'_i)= P(V_i\mid \vec{Z}_i\setminus \{X\},X=x)$ and $P(\vec{Z}'_i=\vec{a},X=x) \ge \alpha^k D_x(\vec{Z}'_i=\vec{a})/|\Sigma|$, the later inequality by marginalization of~\cref{lem:ineqdo} Part 2. over $\vec{V}\setminus(\vec{Z}'_i\cup\{X\})$. Let $N_{i,\vec{a}}$ be the number of samples with with $\vec{Z}'_i=\vec{a}$ if $X\notin \vec{Z}_i$ and with $\vec{Z}'_i\circ X=\vec{a}\circ x$ if $X\in \vec{Z}_i$. Then $N_{i,\vec{a}}\sim \mathrm{Binomial}(m,p)$ where $p \ge \alpha^k D_x(\vec{Z}'_i=\vec{a})/|\Sigma|$.

We partition the summands $(i,\vec{a})$ with $V_i \notin \vec{S}_1$ into two sets: {\em heavy} if $D_x[\vec{Z}'_i=\vec{a}] \ge \epsilon^2/(10n|\Sigma|^{kd+k+1})$ and {\em light} otherwise. Consider the event ``all heavy $(i,\vec{a})$s satisfy $N_{i,\vec{a}} \ge m\alpha^k D_x(\vec{Z}'_i=\vec{a})/(2|\Sigma|)$''. It is easy to see from from our definition of $m$ and heaviness that this event holds except with 1/40 probability from Chernoff and union bounds for a large enough $n$. Hence for the rest of the argument, we condition on this event. In this case, all heavy items satisfy $N_{i,\vec{a}} \ge t$ from our definition of $m$ and $t$.

For the summands $(i,\vec{a})$ with $V_i \notin \vec{S}_1$, we get the following.
\begin{itemize}
\item If $(i,\vec{a})$ is heavy then from~\cref{thm:add1}  $\ex{D_x(\vec{Z}'_i=\vec{a})\; \dkl(D_x(V_i \mid \vec{Z}'_i=\vec{a}),\hat{D}_x(V_i \mid \vec{Z}'_i=\vec{a}))} \le {D_x(\vec{Z}'_i=\vec{a})(|\Sigma|-1) \over N_{i,a}} \le \epsilon^2/(10n|\Sigma|^{kd+k})$, using the lower bound of $N_{i,\vec{a}}$ from the previous paragraph.
\item If a light $(i,\vec{a})$ satisfy $N_{i,\vec{a}} \ge t$, we get $\ex{D_x(\vec{Z}'_i=\vec{a})\; \dkl(D_x(V_i \mid \vec{Z}'_i=\vec{a}),\hat{D}_x(V_i \mid \vec{Z}'_i=\vec{a}))} \le {\epsilon^2\over 10n|\Sigma|^{kd+k+1}}{|\Sigma|-1\over t} \le \epsilon^2/(10n|\Sigma|^{kd+k})$ from~\cref{thm:add1}.
\item $(i,\vec{a})$s which do not satisfy $N_{i,\vec{a}} \ge t$ must be light for which we define the conditional distribution to be uniform. In this case, $\dkl(D_x(V_i \mid \vec{Z}'_i=\vec{a}),\hat{D}_x(V_i \mid \vec{Z}'_i=\vec{a})) = \sum_{\sigma\in \Sigma} D_x(V_i = \sigma \mid \vec{Z}'_i=\vec{a}) \ln (|\Sigma| D_x(V_i \mid \vec{Z}'_i=\vec{a})) = \ln |\Sigma| -H(D_x(V_i \mid \vec{Z}'_i=\vec{a})) \le \ln |\Sigma|$, where $0 \le H(\cdot) \le \ln |\Sigma|$ is the Shannon entropy function. Hence in this case also $\ex{D_x(\vec{Z}'_i=\vec{a})\; \dkl(D_x(V_i \mid \vec{Z}'_i=\vec{a}),\hat{D}_x(V_i \mid \vec{Z}'_i=\vec{a}))} \le \epsilon^2/(10n|\Sigma|^{kd+k})$.
\end{itemize}

Thus each of the $n|\Sigma|^{kd+k}$ summands in the r.h.s. of~\cref{eqn:bnqxadd} is at most $\epsilon^2/(10n|\Sigma|^{kd+k})$ in expectation. We get $\ex{\dkl(D_x,\hat{D}_x)} \le \epsilon^2/10$. From Markov's and Pinsker's inequalities, $\dtv(D_x,\hat{D}_x) \le \epsilon$ except $1/5$ probability.

The total error probability is at most $1/4$ so far.
\end{proof}

We repeat Algorithm~\ref{algo:bn} independently $O(\log {1\over \delta})$ times and use the following result to achieve $(1-\delta)$ success probability.

\begin{theorem}[Theorem 2.9 in~\cite{BGMV20} restated]\label{thm:boost}
Fix any $0<\epsilon,\delta<1$.
Suppose we are given an algorithm that learns an unknown Bayes net $P$ over $\Sigma^N$ on a graph $G$ with indegree $\le \Delta$ as a Bayes net $\hat{P}$ on $G$ such that $\dtv(P,\hat{P})\le \epsilon$ with probability at least 3/4 using $m(\epsilon)$ samples and $t(\epsilon)$ time. Then we can output a distribution $P'$ on $G$ such that $\dtv(P,P')\le \epsilon$ with probability at least $(1-\delta)$ using $O(m(\epsilon/4)\log {1\over \delta})$ samples and $O(t(\epsilon/4)\log {1\over \delta}+|\Sigma|^{2\Delta} N^2\epsilon^{-2}\log^3 {1\over \delta})$ time.
\end{theorem}

We get the following final theorem for learning $P_x$.
\learn*
\ignore{
\begin{restatable}{theorem}{learn}[Evaluation and Sampling]\label{thm:learn}
For any intervention $x$ to $X$ and parameter $\eps \in (0,1)$, there is an algorithm that takes $m = \tilde{O}\left(\frac{|\Sigma|^{2kd}n}{\alpha^k\epsilon^{2}}\right)$ samples from $P$, and in $\tilde{O}\left(\frac{|\Sigma|^{4kd}n^2}{\alpha^k\epsilon^2}\right)$ time, learns a distribution $\hat{P}$ satisfying $\dtv(P_x,\hat{P})\le \epsilon$ such that
\begin{itemize}
\item{Evaluation:} Given an assignment $\vec{w}$ to $\vec{V}\setminus \{X\}$ computing $\hat{P}(\vec{w})$ takes $O(n|\Sigma|(kd+k))$ time 
\item{Sampling:} Obtaining an independent sample from $\hat{P}$ takes $O(n|\Sigma|(kd+k))$ time .
\end{itemize}
\end{restatable}
}
\begin{proof}
We use Algorithm~\ref{algo:bn} which guarantees 3/4 success probability in learning $D_x$ within total variation distance at most $\epsilon$. Then we use~\cref{thm:boost} to improve the success probability to $1-\delta$. The final time and sample complixities follow from~\cref{thm:boost}. 

This gives us a distribution $\hat{D}_x$ over $\vec{V}$, whose marginal distribution on all variables but $X$, we use for evaluation and sampling. Once we have learnt $D_x$, sampling and evaluation takes $O(n|\Sigma|(kd+k))$ time. 
\end{proof}


\section{Lower bound}
\begin{figure*}[t]
\centering
\begin{subfigure}[t]{0.4\textwidth}
\centering
\begin{tikzpicture}
\draw [arrow] (1,1) -- (0,0);
\draw [arrow] (0,0) -- (1,-1);
\draw [arrow] (1,1) -- (1,-1);
\node[text width=0.3cm] at (-0.2,0) {$X$};
\node[text width=0.3cm] at (1,1.2) {$Z$};
\node[text width=0.3cm] at (1,-1.2) {$Y_1$};

\draw [arrow] (1,1) -- (2,-1);
\draw [arrow] (0,0) -- (2,-1);
\node[text width=0.3cm] at (2,-1.2) {$Y_2$};

\node[text width=0.3cm] at (3,-1.2) {$\dots$};

\draw [arrow] (1,1) -- (4,-1);
\draw [arrow] (0,0) -- (4,-1);
\node[text width=0.3cm] at (4,-1.2) {$Y_n$};

\end{tikzpicture}

\caption{With no control variables}
\label{fig:ADMGlower}
\end{subfigure}
~
\begin{subfigure}[t]{0.4\textwidth}
\centering
\begin{tikzpicture}
\draw [arrow] (1,1) -- (0,0);
\draw [arrow] (0,0) -- (1,-1);
\draw [arrow] (1,1) -- (1,-1);
\node[text width=0.3cm] at (-0.2,0) {$X$};
\node[text width=0.3cm] at (1,1.2) {$Z$};
\node[text width=0.3cm] at (1,-1.2) {$Y_1$};

\draw [arrow] (1,1) -- (2,-1);
\draw [arrow] (0,0) -- (2,-1);
\node[text width=0.3cm] at (2,-1.2) {$Y_2$};

\node[text width=0.3cm] at (3,-1.2) {$\dots$};

\draw [arrow] (1,1) -- (4,-1);
\draw [arrow] (0,0) -- (4,-1);
\node[text width=0.3cm] at (4,-1.2) {$Y_n$};

\node[text width=0.3cm] at (3,0.2) {$W_1$};
\node[text width=0.3cm] at (4,0.2) {$\dots$};
\node[text width=0.3cm] at (5,0.2) {$W_d$};

\draw [arrow] (3,0) -- (1,-1);
\draw [arrow] (3,0) -- (2,-1);
\draw [arrow] (3,0) -- (4,-1);

\draw [arrow] (5,0) -- (1,-1);
\draw [arrow] (5,0) -- (2,-1);
\draw [arrow] (5,0) -- (4,-1);

\end{tikzpicture}

\caption{With control variables}
\label{fig:ADMGlowerDegree}
\end{subfigure}
\caption{ADMGs for~\cref{thm:lbmain}}
\label{fig:ADMGlowerboth}
\end{figure*}

For the lower bound we use a well-known packing argument based on Fano's inequality which says if there is a class of $2^{K}$ distributions with pairwise KL distance at most $\beta$ then $\Omega(K/\beta)$ samples are needed to identify a uniformly random distribution from the class. The KL distance is known to satisfy certain chain rule which we use in the following proof (see eg. Lemma 6 in~\cite{lecnote1}).
We first recall \cref{thm:lbmain}.
\lbmain*

\ignore{
\begin{theorem}\label{thm:lower}
Let $G$ be a ADMG of indegree $d$ on a set of $n$ boolean variables $\vec{V}$. Let $X$ be a variable and $\vec{S}$ its c-component. Suppose each variable of $\Pa^+(\vec{S})$ is individually constant-strongly positive whereas $\Pa^+(\vec{S})$ is jointly $\beta$-strongly positive where $\beta\le \alpha$. Let $x$ be a fixed assignment of $X$ and $\epsilon>0$. Then any general algorithm for
learning the distribution $P_x(\vec{V}\setminus \{X\})$ in $\dtv$ distance $\epsilon$ with probability 9/10 requires $\Omega(n2^d/\alpha\epsilon^2)$ samples.
\end{theorem}}
\begin{proof}
We first show the lower bound where $Z$ is a parent of $X$, and $d=2$. Later we show how to prove the full theorem. 

Our ADMG on $n+2$ variables: $Z, X, Y_1, Y_2, \dots, Y_n$ consists of $n$ triangles with $Z,X,Y_j$ for every $j$ where $Z$ is the source and $Y_j$ is the sink. Let $\vec{Y}= \langle Y_1, Y_2, \dots, Y_n\rangle$. Please refer to~\cref{fig:ADMGlower}.

$Z$ is uniform over $\{0,1\}$. $X=\bar{Z}$ with probability $\alpha$ and $X=Z$ with probability $1-\alpha$. Thus $X,Z$ jointly satisfy $\alpha$-strong positivity. Each $Y_j\mid X,Z$ is one among the following two conditional distributions:
\begin{align*}
&D_1: Y_j=\bern(1/2+\epsilon/\sqrt{n})\text{ if }X\neq Z,\\
& \quad\quad Y_j=\bern(1/2)\text{ if }X=Z\\
&D_2: Y_j=\bern(1/2-\epsilon/\sqrt{n})\text{ if }X\neq Z\\
&\quad\quad Y_j=\bern(1/2)\text{ if }X=Z
\end{align*}
We create a class $\mathcal{C}_\epsilon$ of causal models using a code $\mathcal{C}\subset \{0,1\}^n$. This code has size $2^{\Omega(n)}$, and any two of them $c,d\in \mathcal{C}$ satisfy the following: there are $\Theta(n)$ positions where $c$ is 1 and $d$ is 0. Showing existence of such a code is standard. 
Given a code as above, corresponding to every $c\in \mathcal{C}$, we create a product distribution $\vec{Y}\mid X,Z$: the 1 positions of $c$ use the distribution $D_1$ and the 0 positions of $c$ use the distribution $D_2$. Together with the distributions of $X,Z$ this defines a causal Bayes net $\mathcal{P}^c$. 

We first lower-bound the distance between the interventional distributions for any two members $\mathcal{P}^c,\mathcal{P}^d\in \mathcal{C}_\epsilon$. Let $\vec{S}$ be the subset of indices from $[n]$ of size $\Theta(n)$ where $c$ is 1 and $d$ is 0. 
\begin{align*}
\dtv(P^c_{X=1}(\vec{Y}),P^d_{X=1}(\vec{Y}))
& \ge \dtv(P^c_{X=1}(\vec{S}),P^d_{X=1}(\vec{S}))
\end{align*}
With 1/2 probability, when $Z=1$ every dimension of both the distributions are $\bern(1/2)$ and therefore have $\dtv=0$. We focus on the other case when every dimension of $P^c_{X=1}(\vec{S})$ follows $D_1$ and $P^d_{X=1}(\vec{S})$ follows $D_2$.
\begin{align*}
\dtv(P^c_{X=1}(\vec{S}),P^d_{X=1}(\vec{S}))  &= 1/2\cdot \dtv(\bern(1/2+\epsilon/\sqrt{n})^{|\vec{S}|},\bern(1/2-\epsilon/\sqrt{n})^{|\vec{S}|})\\
&\ge 1/2\cdot \dtv(\bern(1/2+\epsilon/\sqrt{n})^{|\vec{S}|},\bern(1/2)^{|\vec{S}|})
\end{align*}
\begin{claim}
$\dtv(\bern(1/2+\epsilon/\sqrt{n})^l,\bern(1/2)^l)\ge \Theta(\epsilon)$ for $l=\Theta(n)$, $l\le n$, $\epsilon \le 1/4$.
\end{claim}
\begin{proof}
{\allowdisplaybreaks
\begin{align*}
\dtv(\bern(1/2+\epsilon/\sqrt{n})^l,\bern(1/2)^l) &=\sum_{i=0}^l {l \choose i} |(1/2+\epsilon/\sqrt{n})^i(1/2-\epsilon/\sqrt{n})^{l-i}-1/2^l|\\
&\ge \sum_{i=0}^{l/2} {l \choose i}2^{-l} (1-(1+2\epsilon/\sqrt{n})^i(1-2\epsilon/\sqrt{n})^{l-i})\\
&\ge \sum_{i=0}^{l/2} {l \choose i}2^{-l} (1-\exp(2\epsilon i/\sqrt{n})\exp(-2\epsilon (l-i)/\sqrt{n}))\\
&\ge \sum_{i=l/2-\sqrt{l}}^{l/2} {l \choose i}2^{-l} (1-\exp(-2\epsilon (l-2i)/\sqrt{n}))\\
&\ge \sum_{i=l/2-\sqrt{l}}^{l/2} {l \choose i}2^{-l} 2\epsilon (l/2-i)/\sqrt{n}\\
&= \sum_{j=0}^{\sqrt{l}} {l \choose l/2-j}2^{-l}2\epsilon j/\sqrt{n}\\
\end{align*}}
The third line in the above uses the fact that in the range $0,1,\dots,l/2$; $1/2^l$ is larger than the other term. The fourth line uses $e^x \ge 1+x$ and $e^{-x}\ge 1-x$. The sixth line uses $1-e^{-x} \ge x/2$ whenever $x\le 1$.  The ratio ${l \choose l/2}/{l \choose l/2-j}$ can be upper-bounded by $\exp(j^2/(l/2-j+1))=O(1)$ for $0\le j \le \sqrt{l}$ and ${l \choose l/2}\simeq 2^l/\sqrt{l}$, which gives $\Theta(\epsilon)$ in the last summation.
\end{proof}

Next we upper bound the KL distance of any two observational distributions from $\mathcal{C}_\epsilon$. Considering the extreme case, we only upper bound the pairs $P^c$ and $P^d$ whose all the coordinates of $\vec{Y}$ are different: one is $D_1$, other is $D_2$. Note that the distributions $P^c|_{X\cup Z}=P^d|_{X\cup Z}:=P(X,Z)$ (say), which gives (by the chain rule): 

\begin{align*}
\mathrm{KL}(P^c,P^d) &= \sum_{x,z} P(x,z)\mathrm{KL}(P^c|_{\vec{Y}},P^d|_{\vec{Y}})\\
&= \sum_{x\neq z} P(x,z) \mathrm{KL}(P^c|_{\vec{Y}},P^d|_{\vec{Y}}) &&\tag{as when $x=z$ both are $\bern(1/2)^n$}\\
&= \alpha \mathrm{KL}(P^c|_{\vec{Y}},P^d|_{\vec{Y}})\\
\end{align*}
$P^c|_{\vec{Y}}$ and $P^d|_{\vec{Y}}$ are product distributions (in the extreme case) whose each component pairs are distributed as either $\bern(1/2+\epsilon/\sqrt{n}),\bern(1/2-\epsilon/\sqrt{n})$ or $\bern(1/2-\epsilon/\sqrt{n}),\bern(1/2+\epsilon/\sqrt{n})$. Using additive property of KL we get $\mathrm{KL}(P^c,P^d)=\Theta(\alpha\epsilon^2)$.

Therefore from Fano's inequality, learning each interventional distribution up to $\Theta(\epsilon)$ distance with probability 2/3 requires $\Omega(n/\alpha\epsilon^2)$ samples.

We next improve the lower bound by $|\Sigma|^d$ factor where $d$ is the indegree. Please refer to~\cref{fig:ADMGlowerDegree}. We pick $|\Sigma|^d$ random models from $\mathcal{C}_\epsilon$ and use them as the conditional distributions for $\vec{Y}\mid X,Z$. Then we create $d$ more control variables $\vec{W}=\langle W_1,\dots,W_d\rangle$ which are uniformly distributed over $\Sigma^d$ and indicates which of the hard distributions is followed by $\vec{Y}$. Any algorithm that want to learn the interventional distribution $X=1$ for this model in $\dtv$ distance $\epsilon$ have to learn a constant fraction of $|\Sigma|^d$ many hard interventional distributions from $\mathcal{C}_\epsilon$ in $\dtv$ distance $O(\epsilon)$  over the choices of $\vec{W}$. We have already established that for every fixing of $\vec{W}$ learning the interventional distributions with 2/3 probability requires $\Omega(n/\alpha\epsilon^2)$ samples. From Chernoff's bound, learning $9\cdot|\Sigma|^d/10$ many distributions with 9/10 probability would require $\Omega(n |\Sigma|^d/\alpha\epsilon^2)$ samples.

We next show how to add hidden variables to the graph. Instead of $Z$ being a parent of $X$ in \cref{fig:ADMGlowerDegree}, suppose that $Z$ is confounded with $X$. That is, there is a hidden variable $U$ that is a parent of $X$ as well as $Z$. Now, we can define the same causal models that we analyzed earlier, with $U$ taking the place of the old $Z$, and the new $Z$ copying the value of $U$. The analysis remains unchanged, as $Z$ is not affected by an intervention on $X$.  
Also, the degree of $X$ and the size of the c-component can be made arbitrarily large by adding dummy variables.
\end{proof}

\section{Evaluation of Marginal Interventions} \label{identifying-marginals}
Here we discuss the problem of estimating $P_x|_{\vec{F}}$, i.e., the marginal interventional distribution of the intervention $x$ to $X$ on a subset of the observables $\vec{F} \subseteq \vec{V}$.  For ease of exposition, we can assume that the vertices of $G$ are $\An^{+}(\vec{F})$ as other variables do not play a role in $P_x|_{\vec{F}}$ and hence can be pruned out from the model; here $\An^+(\vec{F})$ denotes the set of all observable ancestors of $\vec{F}$, including $\vec{F}$.  Tian and Pearl \cite{TP02b} provided an algorithm for this identification question when the ADMG satisfies \cref{ass:id} (See Theorem~4 of \cite{TP02b}), a sufficient condition for identifiability\footnote{Recall that \cite{TP02b} proved~\cref{ass:id} is necessary and sufficient for identifiability of $P_x$.  However to identify $P_x|_{\vec{F}}$, \cref{ass:id} was known to be only sufficient for identifiability.}.  Later works \cite{SP06,HV08} generalized this result of Tian and Pearl for more general interventions, thus exhibiting a sufficient and necessary identifiability graphical condition for this problem.

We consider the following setting: Suppose $\cP$ is an unknown causal Bayes net over a known ADMG $G$ on $n$ observable variables $\vec{V}$ that satisfies (\cref{ass:id}) and $\alpha$-strong positivity with respect to a variable $X \in \vec{V}$ (\cref{ass:bal}) and let $\vec{F} \subseteq \vec{V}$.  Let $d$ denote the maximum in-degree of the graph $G$, $k$ denote the size of its largest c-component, and $f = |\vec{F}|$.  When the graph being referred to is unclear, we will subscript notation (eg: $\Pa_H(V)$ indicates the observable parents of $V$ in graph $H$) to indicate the graph on which the operator is defined on.

We show finite sample bounds for estimating $P_x|_{\vec{F}}$ when the underlying ADMG satisfies~\cref{ass:id}, thus making results of \cite{TP02b} quantitative.  Estimating such causal effects under the necessary and sufficient graphical conditions of \cite{SP06,TP02b} in the finite sample regime is an important and an interesting open question which we leave for future work.  As mentioned in~\cref{contributions}, the required marginal distribution $P_x|_{\vec{F}}$ can be estimated by taking $O(|\Sigma|^{f} / \eps^2)$
samples from the generator $\hat{P}_x$, and we can use \cref{thm:learn} to obtain the generator distribution $\hat{P}_x$ where we require $O( |\Sigma|^{5kd} n / \alpha^k \eps^2)$ many samples from the observational distribution $P$.  Hence we get \cref{cor:learn3}.

The time complexity of the algorithm (of \cref{cor:learn3}) described above is exponential in $f$.  To handle problems that arise in practice for small $\vec{F}$'s, it is of interest to develop efficient algorithms for estimating $P_x|_{\vec{F}}$.  In such cases the approach discussed above is superfluous, as the sample complexity depends linearly on $n$, the total number of variables in the model, which could be unnecessarily large. \cref{thm:learn3}, restated below, shows a sample and time-efficient algorithm when $f$ is very small (e.g. constant).

\learnmarg*

\noindent
The rest of this section is dedicated towards proving \cref{thm:learn3}.

First let us discuss a high level idea of our algorithm for \cref{thm:learn3}.  The idea to handle cases with small $f$ is to restrict our attention to the marginal distribution $P(\vec{W})$ over a small set of vertices $\vec{W}$ and then apply Theorem~3 of \cite{TP02b} (\cref{thm:tp} here) over $\vec{W}$.  However restriction to $\vec{W}$ could potentially modify the parent/child or c-component relationships (or both) across the vertices of $\vec{W}$.  Hence, to apply~\cref{thm:tp}, the underlying causal graph over the vertices of $\vec{W}$ should be obtained via a formal approach in such a way that the topological ordering and the conditional independence relations across vertices of $\vec{W}$ are preserved.  To do that we make use of a well-known latent projection algorithm \cite{verma-pearl} to reduce the given ADMG $G$ (over observables $\vec{V}$) to a different ADMG $H$ (over observables $\vec{W}$).  A similar reduction using the latent projection \cite{verma-pearl} has also appeared in a slightly different context \cite{TK18} -- where the objective is to improve the efficiency of the algorithm.

We carefully choose the set $\vec{W}$ and prune all the other variables $\vec{V} \setminus \vec{W}$ from the graph $G$, and then apply the latent projection to obtain $H$ such that:
\begin{itemize*}
\item[(A)] The required causal effect is identifiable in $H$; 
\item[(B)] \cref{thm:learn} can be applied by maintaining bounds on in-degree and c-component size of this new graph $H$.  
\end{itemize*}

We will show that for $\mv{W} = \vec{F} \cup \Pa^{+}_{G}(\vec{S}_1)$ -- both (A) and (B) hold.  We will prove (A) while we prove (B); although (A) can easily be verified by pruning the vertices of $\vec{V} \setminus \vec{W}$, one by one, by using Corollary~16 of \cite{TK18}.  Before we prove (B) we will first describe the reduction procedure so that the essense of the argument becomes clearer.  Our reduction procedure is discussed next.

The reduction consists of two steps:  The first step is to simplify the given graph $G$ to a much smaller graph $G^{\prime}$ (defined over observables $\vec{W}$) by ignoring all the other variables of $G$ (i.e., ignoring $\vec{V} \setminus \vec{W}$).  By ignoring a certain variable we mean that the variable is considered to be hidden.  Although the observable vertices of $G^{\prime}$ is $\vec{W}$, as desired, $G^{\prime}$ is not an ADMG\footnote{Recall that an ADMG is a graph where the unobservables are root nodes and have exactly two observable children -- denoted by bidirected edges.}.  Since ADMGs are, in general, easy to analyze and parse through we then convert this general causal graph $G^{\prime}$ to an ADMG $H$ using a known reduction technique -- and this is the second step.  This reduction procedure, which we call Reduction($G,\vec{W}$) is formally discussed next.

\subsection{Reduction: Pruning $G$ to a simpler graph $H$}
\paragraph*{Reduction($G,\vec{W}$)}
\begin{enumerate}
\item Let $G^{\prime}$ be the graph obtained from $G$ by considering $\vec{V \setminus W}$ as hidden variables.
\item  \textbf{Projection Algorithm ($G^{\prime}$ to $H$) \cite{TP02a,verma-pearl}}. The projection algorithm reduces the causal graph $G^{\prime}$ to an ADMG $H$ by the following procedure:
\begin{enumerate}
\item For each observable variable $V_i \in \vec{V}$ of $G^{\prime}$, add an observable variable $V_i$ in $H$.
\item For each pair of observable variables $V_i,V_j \in \mv{V}$, if there exists a directed edge from $V_i$ to $V_j$ in $G^{\prime}$, or if there exists a {\em directed} path from $V_i$ to $V_j$ that contains only unobservable variables in $G^{\prime}$, then add a directed edge from $V_i$ to $V_j$ in $H$.  
\item For each pair of observable variables $V_i,V_j \in \mv{V}$, if there exists an unobservable variable $U$ such that there exist two {\em directed} paths in $G^{\prime}$ from $U$ to $V_i$ and from $U$ to $V_j$ such that both the paths contain only unobservable variables, then add a bidirected edge between $V_i$ and $V_j$ in $H$.
\end{enumerate}
\item \textbf{Return} H
\end{enumerate}

\subsection{Properties of Reduction($G,\vec{W}$)}
It is well-known that the projection algorithm ($G^{\prime}$ to $H$) \cite{TP02a,verma-pearl} preserves some of the important properties such as topological ordering and conditional independence relations.  Before we discuss those, let us revisit the equivalent definitions of parents and c-components for general causal graphs with hidden variables.
\begin{definition}[Effective Parents for general causal graphs] \label{def:effParents}
Given a general causal graph $G^{\prime}$ and a vertex $V_j \in \vec{V}$, {\em the effective parents of $V_j$} is the set of all observable vertices $V_i$ such that either $V_i$ is a parent of $V_j$ or there exists a directed path from $V_i$ to $V_j$ that contains only unobservable variables in $G^{\prime}$.
\end{definition}
\begin{definition}[c-component for general causal graphs]
For a given general causal graph $G^{\prime}$, two observable vertices $V_i$ and $V_j$ are related by the {\em c-component relation} if (i) there exists an unobservable variable $U$ such that $G^{\prime}$ contains two paths (a) from $U$ to $V_i$; and (b) from $U$ to $V_j$, where both the paths use only unobservable variables, or (ii) there exists another vertex $V_{z} \in \mv{V}$ such that $V_i$ and $V_{z}$ (and) $V_j$ and $V_{z}$ are related by the c-component relation.
\end{definition}
The below lemma illustrates: ``c-component that contains $X$ remains the same in $G$ and $H$.''
\begin{lemma} \label{lemma:s1_unmodified}
Let $\vec{S}_1$ denotes the c-component that contains $X$ in $G$, $\vec{W} = \vec{Y} \cup \Pa^+_{G}(\vec{S}_1)$ and $H=\text{Reduction}(G, \vec{W})$.  Then, (i) the c-component that contains $X$ in $H$ is also $\vec{S}_1$; (ii) $H$ satisfies~\cref{ass:id}; (iii) $\Pa^+_G(\vec{S}_1)$ and $\Pa^+_{H}(\vec{S}_1)$ are the same; (iv) $H$ satisfies~\cref{ass:bal}.
\end{lemma}
\begin{proof}
Let $\vec{C}$ denote the c-component of $H$ that contains $X$.  Note that $\vec{S}_1 \subseteq \vec{C}$ since all those bidirected edges in $G$ that forms $\vec{S}_1$ are retained in $H$ (because $\vec{S}_1 \subseteq \vec{W}$).  Next we will prove that no other vertex of $H$ share a bidirected edge with $\vec{S}_1$ in H.  Suppose for contradiction there exists a vertex $W_i \in \vec{W}$ that share a bidirected edge with some node $W_j \in \vec{S}_1$ in $H$.  This implies, during the reduction, there exists two paths in $G^{\prime}$ ($U$ to $W_i$ and $U$ to $W_j$) such that all the variables included in these two paths, other than $W_i$ and $W_j$, are unobservables in $G^{\prime}$ which means all those vertices belong to $\vec{V} \setminus \vec{W}$, a contradiction to the fact that $\Pa^+_G(\vec{S}_1)$ is contained in $\vec{W}$. This proves (i).  Since $\Pa^+_G(\vec{S}_1) \subseteq \vec{W}$ there can not exist a bidirected edge between $X$ and a child of $X$ in $H$ which proves (ii).  

Note that $\Pa^{+}_{G}(\vec{S}_1) \subseteq \Pa^{+}_{H}(\vec{S}_1)$ -- because $\vec{W}$ contains $\Pa^{+}_{G}(\vec{S}_1)$.  Now suppose, for contradiction, $\Pa^{+}_{G}(\vec{S}_1) \subset \Pa^{+}_{H}(\vec{S}_1)$. Then during the reduction step there must have been an edge from an unobservable to $\vec{S}_1$ in $G^{\prime}$, which can not be true as $\vec{W}$ contains $\Pa^{+}_{G}(\vec{S}_1)$ and none of those variables are treated as hidden variables in the reduction.  This proves (iii).  Since $\vec{S}_1$ and $\Pa^{+}(\vec{S}_1)$ remains unchanged in both $G$ and $H$,~\cref{ass:bal} still holds in $H$ which proves (iii). 
\end{proof}
The projection algorithm ($G^{\prime}$ to $H$) is known to preserve the following set of properties.
\begin{itemize}
\item The $c$-components of $H$ and $G^{\prime}$ are identical and the c-component factorization formula (Equation~($20$) in Lemma~$2$ of \cite{TP02a}) holds even for the general causal graph (See Section~5 of \cite{TP02a} for more details).  They show this based on a known previously known reduction from $G^{\prime}$ to $H$ \cite{verma-pearl}.  The proof is based on the fact that for any subset $\mv{S} \subseteq \mv{V}$ of observable variables, the induced subgraphs $G^{\prime}[\mv{S}]$ and $H[\mv{S}]$ require the same set of conditional independence constraints.
\item The effective parents (see~\cref{def:effParents}) of every observable node in $G^{\prime}$ is the same as the (observable) parent set of the corresponding node in $H$.  
\item The observable vertices of $G^{\prime}$ and $H$ are the same.  
\item Also, the topological ordering of the observable nodes of $G^{\prime}$ and $H$ are the same.
\end{itemize}
As is common in the causality literature we do not use any other property of $G^{\prime}$ besides the above in our analysis, and hence it is sufficient to derive conclusions from this modified graph which contains only a small number of vertices.  
\subsubsection{Proof of \cref{thm:learn3}}
We know from~\cref{lemma:s1_unmodified} that whenever $G$ satisfy~\cref{ass:id}, $H= \text{Reduction}(G,\vec{W})$ with $\vec{W}= \vec{Y} \cup \Pa^+_{G}(\vec{S}_1)$ satisfy~\cref{ass:id} as well.  For such graphs $G$ and $H$, while both satisfy~\cref{ass:id}, it is well-known that an equivalent statement of~\cref{thm:tp} directly follows from the proof of Theorem 3 of \cite{TP02b} since their proof uses only the above mentioned properties.  We will also extensively use (i) of \cref{lemma:s1_unmodified}.  This results in the following theorem.
\begin{theorem}[Theorem 3 of \cite{TP02b} with respect to ADMG $H = \text{Reduction}(G,\vec{W})$]\label{thm:tp_marginal}
Let $P$ be a CBN over a causal graph $G=(\vec{V},E^\to \cup E^\leftrightarrow)$, $X\in \vec{V}$ be a designated variable and $\vec{Y} \subseteq \vec{V \setminus X}$.  Let $\vec{S}_1, \dots, \vec{S}_{\ell}$ are the c-components of $G$ and without loss of generality assume $X\in \vec{S}_1$. Suppose that $G$ satisfies \cref{ass:id}.  Let $\vec{W} = \vec{Y} \cup \Pa^+_G(\vec{S}_1)$ for $\vec{Y} \subseteq \vec{V}\setminus\{X\}$.  Let $H =\text{Reduction}(G,\vec{W})$ and let $\vec{S}^{\prime}_1, \dots, \vec{S}^{\prime}_{\ell'}$ be the c-components of $H$ where without loss of generality let $\vec{S}_1 = \vec{S}^{\prime}_1$.  Then for any setting $x$ to $X$ and any assignment $\vec{t}$ to $\vec{W}\setminus \{X\}$ the interventional  distribution $P_x(\vec{t})$ is given by: 
\begin{align*}
P_x(\vec{t}) &= P_{\vec{t}_{\vec{W}\setminus \vec{S}^{\prime}_1}}(\vec{t}_{\vec{S}^{\prime}_1\setminus \{X\}}) \cdot \prod_{j=2}^{\ell'} P_{\vec{t}_{\vec{W}\setminus (\vec{S}^{\prime}_j\cup \{X\})}\circ x}(\vec{t}_{\vec{S}^{\prime}_j}) = \sum_{\tilde{x} \in \Sigma}Q_{\vec{S}^{\prime}_1}(\vec{t}\circ \tilde{x}) \cdot \prod_{j=2}^{\ell'} Q_{\vec{S}^{\prime}_j}(\vec{t}\circ x)
\end{align*}
\end{theorem}
This proves part (A) discussed before.  Next we prove part (B): where we provide bounds on the in-degree and the cardinality of c-components of $H$.
\begin{lemma}
Let $\vec{S}_1$ be the c-component of $G$ that contains $X$.  Let $\vec{W}= \vec{Y} \cup \Pa^+_G(\vec{S}_1)$, $H = \text{Reduction}(G,\vec{W})$ and let the c-components of $H$ are denoted by $\vec{S}^{\prime}_1, \dots, \vec{S}^{\prime}_{\ell^{\prime}}$ without loss of generality let $\vec{S}_1 = \vec{S}^{\prime}_1$.  Then:
\begin{enumerate}
\item The in-degree of $H$ is at most $f + k(d+1)$.
\item $\lvert \vec{S}^{\prime}_i \rvert \leq f + kd$, for every $i$.
\end{enumerate}
\end{lemma}
\begin{proof}
The fact that $H$ contains at most $f + k(d+1)$ vertices provides the bound on the in-degree.  We know from~\cref{lemma:s1_unmodified} that $\vec{S}^{\prime}_1$ is a c-component of $H$ and the remaining vertices of $H$ is $(\vec{Y} \cup \Pa^{+}_{H}(\vec{S}^{\prime}_1)) \setminus \vec{S}^{\prime}_1$ which is of size at most $f + k d$ (since $\vec{S}_1 = \vec{S}_1^{\prime}$ and $\Pa_G^+(\vec{S}_1) = \Pa_H(\vec{S}_1)$).
\end{proof}
We have now gathered the tools required to prove~\cref{thm:learn3}.
\begin{proof}[Proof of~\cref{thm:learn3}]
Let $\vec{W} = \vec{Y} \cup \Pa^{+}_{G}(\vec{S}_1)$ and let $H = \text{Reduction}(G,\vec{W})$.  The reduction $H = \text{Reduction}(G,\vec{W})$ can be performed using breadth first search/depth first search which can be done in time linear in the size of the input graph.  We obtained an equivalent statement of~\cref{thm:tp} in~\cref{thm:tp_marginal}.  Also from~\cref{lemma:s1_unmodified} we know that the model over the causal graph $H$ satisfies both~\cref{ass:id} and \cref{ass:bal}.  Hence by substituting $n$ by $f + k(d+1)$ -- the cardinality of observables of $H$; $k$ by $f + kd$ -- the size of the largest c-component of $H$; and $d$ by $f + k(d+1)$ -- the in-degree of $H$, into~\cref{thm:learn} we obtain the desired bounds on sample and time complexities.
\end{proof}

\bibliographystyle{alpha}
\bibliography{reference} 
\end{document}